\documentclass{article} % For LaTeX2e

\usepackage{iclr2027_conference,times}

\iclrfinalcopy % Uncomment for camera-ready version, but NOT for submission.
\usepackage[utf8]{inputenc} % allow utf-8 input
\usepackage[T1]{fontenc}    % use 8-bit T1 fonts
\usepackage{hyperref}       % hyperlinks
\usepackage{url}            % simple URL typesetting
\usepackage{booktabs}       % professional-quality tables
\usepackage{amsfonts}       % blackboard math symbols
\usepackage{nicefrac}       % compact symbols for 1/2, etc.
\usepackage{microtype}      % microtypography
\usepackage{xcolor}         % colors
\usepackage{multirow} % 处理跨行
\usepackage{graphicx} % 提供 \resizebox 用于调整表格宽度
\usepackage{amsmath}
\usepackage{multirow}
\usepackage{booktabs}
\usepackage[most]{tcolorbox} % 提供带背景的 prompt 方框
\author{Shuyu Guo \\
  Shandong University \\
  Qingdao, China \\
  \texttt{guoshuyu225@gmail.com} \\\And
  Shuo Zhang \\
  Bloomberg \\
  London, United Kingdom \\
  \texttt{szhang611@bloomberg.net} \\\And
  Zhaochun Ren\thanks{Corresponding Author.} \\
  Leiden University \\
  Leiden, The Netherlands \\
  \texttt{z.ren@liacs.leidenuniv.nl} \\}

\title{Compression Beyond the Uncompressed: A Two-Stage Training Recipe for Soft Context Compression in RAG}

\begin{document}

\maketitle
\lhead{Preprint}

\begin{abstract}
Retrieval-Augmented Generation (RAG) improves knowledge-intensive generation by conditioning language models on retrieved documents, but processing these documents becomes increasingly expensive as retrieval depth grows.
Soft context compression reduces this cost by encoding documents into compact continuous representations that can be precomputed and reused across queries.
However, many existing methods train compressed models by distilling from a full-context teacher.
When the teacher is wrong, such distillation can reinforce its errors, while teacher imitation provides no direct signal for improving beyond the teacher.
We propose DEX-Comp, a two-stage training recipe that separates reliable imitation from targeted exploration.
Pure Distillation learns only from teacher-correct questions to mitigate error propagation, while Hard Exploration applies outcome-based reinforcement learning to teacher-failed questions to directly optimize answer correctness.
Across five open-domain QA benchmarks and retrieval depths from top-$5$ to top-$30$, DEX-Comp at $16\times$ compression outperforms all evaluated compression baselines and surpasses the untuned full-context RAG model in average accuracy, while reducing time-to-first-token by $4.4\times$--$23.7\times$.
Evaluations across additional datasets and backbones further demonstrate its generalization.

% Retrieval-augmented generation (RAG) enhances language models with external knowledge, but lengthy retrieved contexts increase inference cost.
% % SZ: please use Retrieval-Augmented Generation (RAG) everywhere
% % SZ: Lengthy?
% Soft context compression alleviates this cost by encoding documents into shorter embedding sequences.
% % SZ: What does soft mean
% Although distillation-based training is simple and effective, unfiltered response distillation can reinforce errors from the full-context teacher, while optimizing teacher agreement provides no direct reward for improving answer correctness under compressed inputs.
% % SZ: distillation-based training, unfiltered response distillation etc, under defined
% To address these limitations, we propose DEX-Comp, a two-stage training recipe:
% Pure Distillation learns only from correct teacher responses to inherit answering ability while mitigating error reinforcement.
% Hard Exploration then applies reinforcement learning to teacher-failed questions, directly rewarding correct answers from compressed inputs.
% Across five open-domain QA benchmarks at retrieval depths from top-$5$ to top-$30$, DEX-Comp outperforms all evaluated compression baselines and exceeds the full-context model in average accuracy, while compressing inputs by $16\times$ and accelerates inference by $4\times$--$24\times$.
% Ablations and evaluations across additional datasets and backbones support the effectiveness of both stages and the generalization of our approach.
\end{abstract}

\section{Introduction}
\label{sec:intro}
Retrieval-Augmented Generation (RAG) improves knowledge-intensive generation by conditioning large language models (LLMs) on retrieved documents~\citep{DBLP:journals/corr/abs-2312-10997}.
% As retrieval depth grows, however, processing retrieved documents in full becomes increasingly expensive in computation and memory~\citep{autocompr}.
% Moreover, retrieval may introduce redundant or irrelevant information that distracts the model and degrades generation quality~\citep{longllmlingua}.
However, expanding the retrieved context introduces two challenges:
(1) Longer inputs increase computation and memory requirements, limiting the efficiency of generation~\citep{autocompr};
and (2) retrieval may introduce redundant or irrelevant information that distracts the model and degrades generation quality~\citep{longllmlingua}.
These challenges motivate context compression: reducing the retrieved context before it is consumed by the language model.

% Existing context compression methods broadly operate in either the discrete token space or the continuous representation space.
Existing approaches broadly fall into hard context compression~\citep{IterCOMP,poc,ShiMNLJ26Discovering,XProvence,TangXLZZHZ25PerceptionCompressor,dac,EfficientPrompt,DynamicCompressing} and soft context compression~\citep{BeyondPos,ReadAsHuman,tang2026comi,tang2026gmsa,ZhaoLTLCZYXZSZ26CoMeT,DBLP:journals/corr/abs-2602-13980,OSCAR,DBLP:conf/emnlp/ZhaoLLHXXZ25,LiSC25500xCompressor}.
Hard compression reduces context by extracting salient spans~\citep{LLMLingua} or summarizing documents into shorter text~\citep{DBLP:conf/acl/JinLDZZWLQD25}, but aggressive compression can discard information useful for downstream generation~\citep{Understanding}.
Soft compression instead encodes each document into a much shorter sequence of continuous embeddings~\citep{Beacon}.
Because these embeddings need not correspond to readable tokens, they provide a flexible learned representation of the original document~\citep{cramming}.
In this work, we focus on \emph{query-independent soft compression}, where documents are compressed without access to the query~\citep{autoencoding,pcc}.
Their compressed representations can therefore be precomputed offline and reused across queries, substantially reducing the context processed by the decoder at inference time.

A central challenge in soft compression is how to train these compressed representations.
A common pipeline first pretrains the compressor on unlabeled text through reconstruction or next-token prediction and then fine-tunes it for downstream RAG question answering (QA)~\citep{icae,cocom}.
For QA fine-tuning, distillation from an uncompressed full-context model has been shown to outperform direct supervision with reference answers, possibly because short reference answers are poorly aligned with the model's response distribution~\citep{xrag,ACC-RAG}.
More recently, \citet{pisco} showed that distillation alone can achieve strong performance without separate compression pretraining.
Together, these results establish the full-context model as an effective source of supervision for learning soft compressed representations.

% Soft compression typically follows a two-stage training pipeline: 
% pretraining on unlabeled text through reconstruction or next-token prediction, followed by fine-tuning for RAG question answering (QA)~\citep{icae,cocom}.
% For QA fine-tuning, prior studies show that distillation from the uncompressed model is more effective than supervision with reference answers, possibly because short reference answers are misaligned with the model's verbose response style~\citep{xrag,ACC-RAG}.
% \citet{pisco} further simplifies this pipeline, showing that distillation alone achieves strong performance without compression pretraining.
% Together, these findings suggest that preserving the uncompressed model's answering behavior is key to effective compression training.

\begin{figure}[t]
 \centering
 \includegraphics[width=\linewidth]{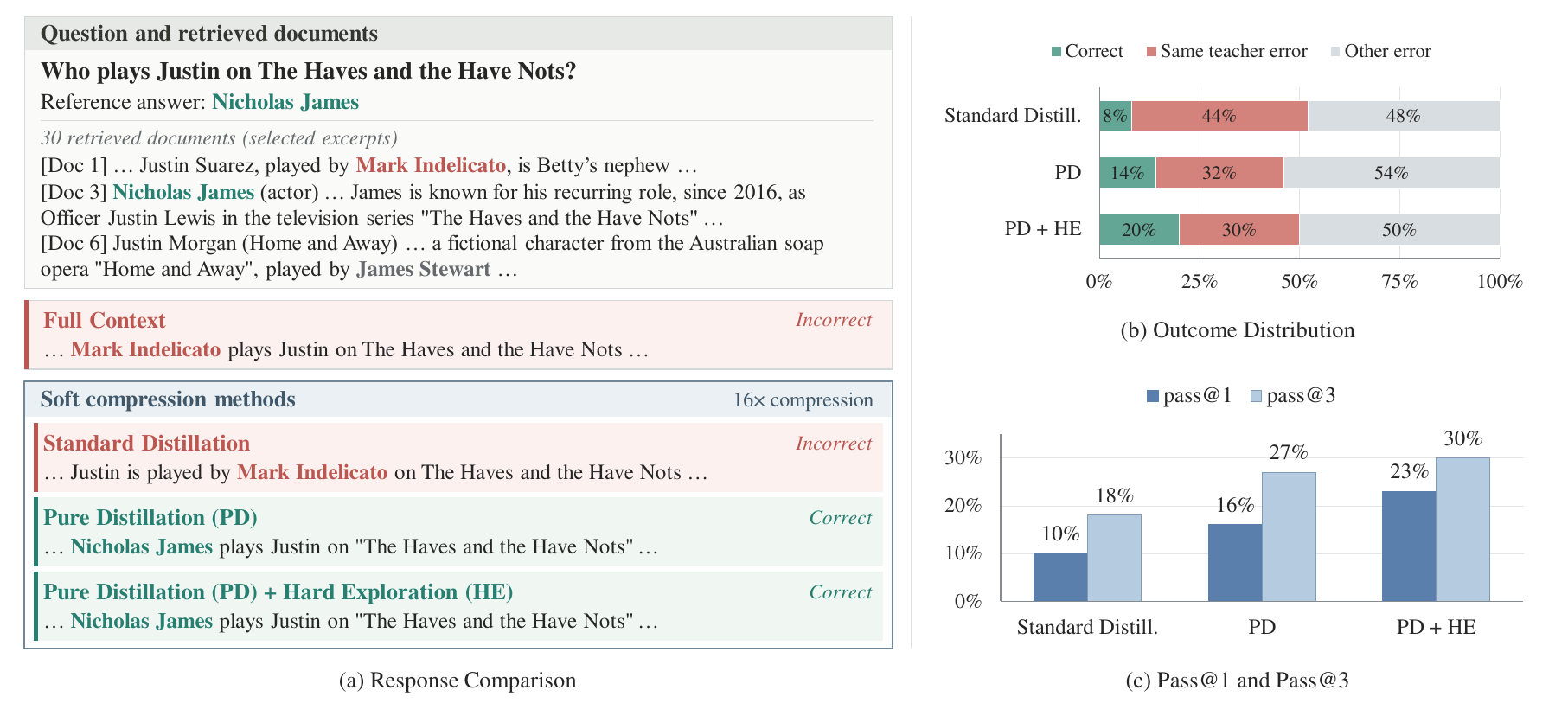}
 \caption{
  \textbf{(a) Representative case.} Standard distillation reproduces an incorrect answer from the full-context teacher, whereas Pure Distillation (PD) and PD with Hard Exploration (PD+HE) answer correctly.
  \textbf{(b) Outcome distribution.} Under greedy decoding on a fixed subset of teacher-failed questions, PD reduces the frequency of reproducing teacher errors, while HE further increases correct answers.
  \textbf{(c) Sampling performance.} On the same questions, PD improves pass@$1$ and pass@$3$ over standard distillation, with further gains from HE.
 }
 \label{fig:distillation_limits}
\end{figure}

However, an effective teacher is not necessarily a reliable oracle.
This creates two limitations for distillation-based compression training.
(1) \emph{Imitation of teacher errors.}
Standard distillation approaches imitate the full-context teacher regardless of whether its response is correct.
When the teacher fails, this supervision can propagate the same error to the compressed model.
Figure~\ref{fig:distillation_limits}(a) shows a representative example: despite the correct evidence appearing in the retrieved documents, the full-context teacher produces an incorrect answer, which is subsequently reproduced by standard distillation.
More broadly, Figure~\ref{fig:distillation_limits}(b) shows that a substantial fraction of errors made by the distilled model match those of the teacher.
(2) \emph{Limited improvement beyond imitation.}
Distillation optimizes agreement with the teacher rather than answer correctness itself.
It therefore provides no direct learning signal for correcting cases in which the teacher fails, placing a natural ceiling on what can be achieved through imitation alone.

% However, two limitations exist in such distillation-based training paradigm:
% (1) Imitation of teacher errors.
% All-response distillation treats every response from the full-context teacher as a learning target, regardless of its correctness.
% % SZ: "All-reponse" is not very precise; Non-filtering?
% Matching an incorrect response can encourage the compressor to preserve distracting information that supports the teacher's mistake, reinforcing the same error in the compressed model.
% % SZ: Matching is not previse
% Figure~\ref{fig:distillation_limits}(a,b) illustrates this risk through a shared incorrect answer and a high rate of teacher-matching errors under all-response distillation.
% % SZ: error propogation?
% (2) Limited improvement beyond imitation.
% Distillation optimizes agreement with the teacher without directly rewarding corrections to its errors.
% The compressed model is thus trained to imitate the teacher's responses to full-context tokens, even though it answers from dense embeddings with greater information capacity per slot~\citep{cramming}.
% This reliance on teacher behavior limits further improvement in compressed models. 
% In our evaluation, all soft compression baselines still fall short of the full-context model (see Table~\ref{tab:results-comparison}).
% These limitations naturally raise two research questions:
% \emph{(1) How can we initialize a compressed model to inherit the full-context model's answering ability without reinforcing its errors?
% (2) How can we further improve a compressed model beyond imitation to close the compression gap?}

This observation suggests that teacher-correct and teacher-failed examples should play different roles in compression training.
When the full-context teacher answers correctly, its behavior provides useful supervision for learning how to answer from compressed representations.
When it answers incorrectly, however, imitating that behavior is undesirable; these examples instead provide an opportunity to optimize the compressed model directly against task outcomes.
This naturally raises two research questions:
\emph{(1) How can we use a full-context teacher to reliably initialize a compressed model without propagating its observed errors?
% (2) How can we exploit teacher-failed examples to improve the compressed model beyond imitation?
(2) How can we directly optimize the compressed model for answer correctness beyond imitation?}

% \begin{figure}[t]
%  \centering
%  \includegraphics[width=\linewidth]{figures/figure1_academic_editable.pdf}
%  \caption{
%   \textbf{(a) Representative case.} All-response distillation (Full distill.) repeats the full-context teacher's error, whereas Pure Distillation (PD) and PD with Hard Exploration (PD+HE) answer correctly.
%  \textbf{(b) Outcome distribution.} Under greedy decoding on a fixed subset of teacher-failed questions, all-response distillation frequently repeats teacher errors; PD reduces this rate, and HE further improves accuracy.
%  \textbf{(c) Sampling performance.} On the same subset, PD improves pass@$1$ and pass@$3$ over all-response distillation, with further gains from HE.
% }
%  \label{fig:distillation_limits}
% \end{figure}

To address these questions, we propose \textbf{DEX-Comp} (\textbf{D}istillation and \textbf{Ex}ploration for soft \textbf{Comp}ression), a two-stage training recipe that assigns different learning signals according to teacher correctness.
In the first stage, \emph{Pure Distillation (PD)}, we distill only on teacher-correct questions, providing a reliable initialization while avoiding direct imitation of the teacher's observed errors (see Figure~\ref{fig:distillation_limits}(b,c)).
In the second stage, \emph{Hard Exploration (HE)}, we apply outcome-based reinforcement learning to teacher-failed questions, rewarding answer correctness rather than agreement with the teacher.
HE therefore turns examples with unreliable imitation targets into opportunities for direct task optimization under compressed inputs, enabling further improvement beyond teacher imitation (see Figure~\ref{fig:distillation_limits}(b,c) and Section~\ref{sec:ablation}).

This separation is particularly attractive for soft compression.
Rather than performing reinforcement learning over the full retrieved context, DEX-Comp performs reinforcement learning over substantially shorter compressed representations.
As a result, it improves over an untuned full-context RAG model at substantially lower training cost than full-context RL and remains feasible at retrieval depths where full-context RL exceeds the available memory budget (Section~\ref{sec:training_cost}).
Thus, compression serves not only as an inference-time efficiency mechanism, but also as an efficient substrate for post-training.

% Our contributions are threefold:
% (1) We identify two limitations of treating the full-context model as an oracle for soft-compression training: indiscriminate distillation can propagate teacher errors, while imitation alone provides no direct mechanism for correcting them.
% (2) We introduce DEX-Comp, which separates reliable imitation from targeted exploration by distilling teacher-correct examples and applying outcome-based reinforcement learning to teacher-failed examples.
% (3) Across five open-domain QA benchmarks and retrieval depths from top-$5$ to top-$30$, DEX-Comp at $16\times$ compression outperforms all evaluated compression baselines and surpasses the untuned full-context RAG model in average accuracy, while reducing time-to-first-token by $4.4\times$--$23.7\times$.
% Additional evaluations across datasets and model backbones further demonstrate its generalization.
Our contributions are threefold:
(1) We identify two limitations of indiscriminate teacher imitation for soft-compression training: distillation can propagate teacher errors, while imitation alone provides no direct mechanism for correcting them.
(2) We introduce DEX-Comp, which assigns different learning signals according to teacher correctness, using reliable distillation on teacher-correct examples and outcome-based reinforcement learning on teacher-failed examples.
(3) Across five open-domain QA benchmarks and retrieval depths from top-$5$ to top-$30$, DEX-Comp at $16\times$ compression outperforms all evaluated compression baselines and surpasses the untuned full-context RAG model in average accuracy, while reducing time-to-first-token by $4.4\times$--$23.7\times$.
Additional evaluations across datasets and model backbones further demonstrate its generalization.

\section{Related Work}
\label{sec:related}

\paragraph{Retrieval-Augmented Generation}
RAG augments language models with external knowledge and has been widely studied for knowledge-intensive tasks~\citep{guan2026deeprag, wei2025instructrag, gutierrez2025HippoRAG, DBLP:conf/naacl/ShiMYS0LZY24, DBLP:conf/acl/MinSL0YHZ23}.
Prior work improves RAG through joint retriever--generator training, retriever alignment, and adaptive retrieval and generation~\citep{DBLP:conf/icml/GuuLTPC20, DBLP:conf/iclr/Lin0CSL00KSLZY24, selfrag}, while a complementary line of work studies the efficiency of processing increasingly large retrieved contexts.

% \paragraph{Context Compression}
% Mitigating the efficiency degradation of RAG can be approached by reducing the decoder's effective computation, broadly along two lines: inference-time KV compression~\citep{deltakv,OmniKV,SnapKV,Palu} and pre-inference context compression~\citep{ccs}.
% The latter reduces the context presented to the decoder and can be divided into hard and soft variants.
% Hard compression operates on the surface form, either by extracting salient tokens~\citep{Leveraging,DChungCLH0Y24Selection-p,EXIT,Provence,0001DGL23SelectContext,DBLP:journals/corr/abs-2304-12102} or by summarizing context~\citep{recomp,discomp,attentionrag,DBLP:conf/acl/JinLDZZWLQD25,LearningtoCompress}.
% Soft compression instead encodes the context into a sequence of continuous embeddings and has shown strong performance relative to hard compression~\citep{AgentOCR}.
% In this work, we focus on query-independent context compression, where documents are encoded offline without access to the query.
% This setting differs from query-dependent compression methods, which select or compress content conditioned on the query~\citep{Rethinking, DAST,DBLP:conf/aaai/ZhaoWX25,DBLP:conf/aaai/LiskavetsURKEL25,CORE,filc,compact}.

\paragraph{Context Compression}
RAG efficiency can be improved through inference-time KV compression~\citep{deltakv,OmniKV,SnapKV,Palu} or pre-inference context compression~\citep{ccs}.
% Hard compression operates on the surface form, either by extracting salient tokens~\citep{Leveraging,DChungCLH0Y24Selection-p,EXIT,Provence,0001DGL23SelectContext,DBLP:journals/corr/abs-2304-12102} or by summarizing context~\citep{recomp,discomp,attentionrag,DBLP:conf/acl/JinLDZZWLQD25,LearningtoCompress}.
The latter includes hard compression, which extracts or summarizes textual content~\citep{Leveraging,DChungCLH0Y24Selection-p,EXIT,Provence,0001DGL23SelectContext,DBLP:journals/corr/abs-2304-12102,recomp,discomp,attentionrag,DBLP:conf/acl/JinLDZZWLQD25,LearningtoCompress}, and soft compression, which encodes context into continuous embeddings~\citep{AgentOCR}.
We focus on query-independent soft compression, where documents are encoded offline without access to the query, unlike query-dependent methods that condition compression on the query~\citep{Rethinking,DBLP:conf/aaai/LiskavetsURKEL25,DAST,CORE,filc,compact}.

% A common training pipeline for soft compression first pretrains a compressor on unlabeled text, then fine-tunes it on downstream QA with reference-answer supervision~\citep{autoencoding, pcc, sara, rmt,ftr,dodo,gist,LLoCO}.
% Some methods replace this supervision with distillation from the uncompressed model~\citep{ACC-RAG, Simple, xrag, YenG024Long-ContextLanguage}.
% \citet{pisco} simplifies the pipeline to distillation alone, removing separate pretraining.
% \citet{ting2026bridging} extend pretraining and reference-answer fine-tuning with reinforcement learning, yielding a three-stage training procedure.
% Existing methods therefore employ different combinations of pretraining, teacher imitation, and outcome optimization, but do not explicitly distinguish when teacher imitation provides reliable supervision and when direct outcome optimization is preferable.
% Our work makes this distinction explicit by using teacher-correct examples for reliable distillation and teacher-failed examples for targeted exploration.
A common soft-compression pipeline first pretrains a compressor on unlabeled text and then fine-tunes it on downstream QA with reference-answer supervision~\citep{sara,rmt,ftr,dodo,gist,LLoCO}.
Other methods use distillation from the uncompressed model~\citep{ACC-RAG,Simple,xrag,YenG024Long-ContextLanguage}, with \citet{pisco} showing that distillation alone can remove separate pretraining.
\citet{ting2026bridging} instead combine pretraining and reference-answer fine-tuning with reinforcement learning.
\emph{Existing methods therefore combine pretraining, teacher imitation, and outcome optimization in different ways, but do not explicitly distinguish when teacher imitation is reliable and when direct outcome optimization is preferable.
DEX-Comp makes this distinction explicit by using teacher-correct examples for distillation and teacher-failed examples for exploration.}
\section{Method}
\label{sec:method}

\paragraph{Problem Formulation.}
A standard RAG system consists of a retriever $\mathcal{R}$ and a decoder $\mathcal{T}$~\citep{DBLP:conf/emnlp/KarpukhinOMLWEC20}.
Given a query $q$, the retriever returns the top-$k$ documents
$\mathbf{D}_q = \{d_1,\ldots,d_k\}$ from a corpus $\mathbb{C}$,
and the decoder generates a response conditioned on the retrieved documents:
\[
r \sim \mathcal{T}(\cdot \mid \mathbf{D}_q, q).
\]

Processing the retrieved documents in full becomes increasingly expensive as retrieval depth grows.
Soft context compression reduces this cost by using a compressor $\mathcal{C}_\phi$ to replace each document $d_i$ with a shorter sequence of continuous embeddings
$\mathbf{e}_i = \mathcal{C}_\phi(d_i)$,
which are consumed by a compression-aware decoder $\mathcal{G}_\theta$.
These continuous representations have no explicit supervision target, making it non-trivial to jointly train the compressor and decoder.
Prior work therefore commonly uses an uncompressed full-context model as a teacher~\citep{xrag,pisco}.
As discussed in Section~\ref{sec:intro}, such a teacher provides useful supervision when it answers correctly, but becomes an unreliable imitation target when it fails.

Motivated by this distinction, we propose \textbf{DEX-Comp}
(\textbf{D}istillation and \textbf{Ex}ploration for soft \textbf{Comp}ression),
a two-stage training recipe that assigns different learning signals according to teacher correctness.
\emph{Pure Distillation (PD)} uses teacher-correct examples to provide a reliable initialization,
while \emph{Hard Exploration (HE)} applies outcome-based reinforcement learning to teacher-failed examples to directly optimize answer correctness.
We first describe the compression architecture (\S\ref{sec:arch}) and then introduce the two training stages
(\S\ref{sec:pd}--\S\ref{sec:he}).

\subsection{DEX-Comp Architecture}
\label{sec:arch}

DEX-Comp consists of a compressor $\mathcal{C}_\phi$ and a decoder $\mathcal{G}_\theta$.
For each document $d_i$, the input text is concatenated with
$m_i = \lfloor L_i / \tau \rfloor$ special compression tokens,
where $L_i$ is the document length and $\tau$ is a fixed compression rate.
The full sequence is encoded by $\mathcal{C}_\phi$, and the final-layer hidden states at the compression-token positions form the compressed embedding sequence
\[
\mathbf{e}_i = \mathcal{C}_\phi(d_i) = \big(e_i^1, \ldots, e_i^{m_i}\big).
\]
The entire compression process is run offline once per document in the corpus $\mathbb{C}$, with all $\mathbf{e}_i$ pre-computed and cached for instant access at inference time.
The decoder is an autoregressive language model.
Given the cached embeddings $\{\mathbf{e}_1, \ldots, \mathbf{e}_k\}$ of the top-$k$ retrieved documents and a query $q$, $\mathcal{G}_\theta$ generates the response token by token from
\[
P_\theta(r_t \mid r_{<t}, q, \mathbf{e}_1, \ldots, \mathbf{e}_k).
\]
% Training $(\phi, \theta)$ jointly thus raises two questions: (1) how to cold-start the coupled system when neither $\mathcal{C}_\phi$ knows what to write into the soft slots nor $\mathcal{G}_\theta$ knows how to read them; and (2) how to adapt this initialization toward a computation strategy better suited to compressed representations.
% The two subsections that follow address them in turn.

\begin{figure}[t]
    \centering
    \includegraphics[width=\linewidth]{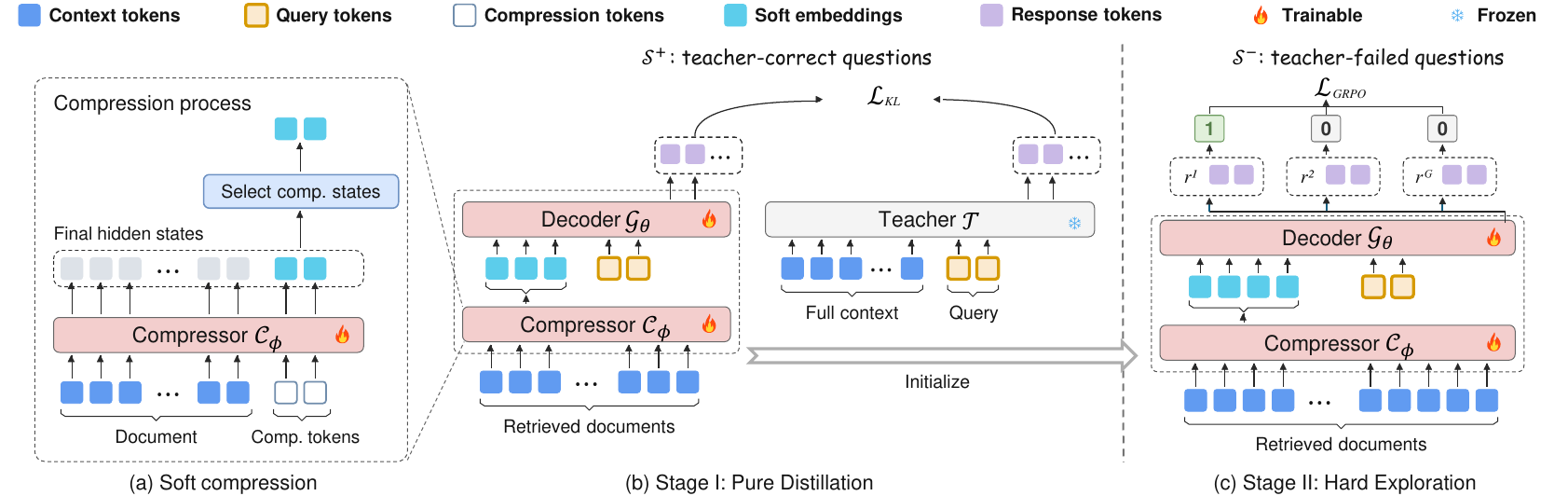}
    \caption{Overview of DEX-Comp.
    \textbf{(a) Soft compression.} A compressor turns each document into a shorter sequence of soft embeddings by selecting the final-layer hidden states at compression-token positions. The embeddings can be precomputed offline and consumed by the decoder at inference time.
    \textbf{(b) Pure Distillation.} On teacher-correct queries ($\mathcal{S}^{+}$), KL distillation aligns the compressed model's response distribution with that of a frozen RAG teacher reading the full retrieved documents.
    \textbf{(c) Hard Exploration.}
    Initialized from Stage~I, the compressed model trains on teacher-failed queries
    ($\mathcal{S}^{-}$) using outcome-based GRPO, directly rewarding answer correctness rather than teacher agreement.}
    \label{fig:method}
\end{figure}

\subsection{Training Stage I: Pure Distillation}
\label{sec:pd}

Distillation provides an effective way to initialize the jointly trained compressor and decoder, but the full-context teacher is not correct on every training example.
We therefore restrict imitation to examples for which the teacher provides a reliable demonstration.
Specifically, we partition the training set according to whether the teacher's response matches the gold answer:
% Cold-starting a jointly optimized $(\mathcal{C}_\phi, \mathcal{G}_\theta)$ is non-trivial.
% Since no oracle embedding is known a priori, we resort to end-to-end distillation from the uncompressed teacher $\mathcal{T}$ as in prior work~\cite{xrag,pisco}.
% In contrast, we distill only the examples that the teacher already answers correctly, and refer to this stage as Pure Distillation (PD).
% Specifically, we partition the training set by whether the teacher's response matches the gold answer:
\begin{equation}
    \mathcal{S}^{+} = \big\{(q, \mathbf{D}_q) : \mathrm{eval}(\hat r_{\mathcal{T}}, r^\star) = 1\big\},
    \qquad
    \mathcal{S}^{-} = \big\{(q, \mathbf{D}_q) : \mathrm{eval}(\hat r_{\mathcal{T}}, r^\star) = 0\big\},
    \label{eq:partition}
\end{equation}
% where $\hat r_{\mathcal{T}} \sim \mathcal{T}(\cdot \mid \mathbf{D}_q, q)$ is the teacher's response and $\mathrm{eval}(\cdot,\cdot) \in \{0,1\}$ checks whether the prediction matches the gold answer $r^\star$.
% where $\hat r_{\mathcal{T}} \sim \mathcal{T}(\cdot \mid \mathbf{D}_q,q)$ is the teacher response,
% $r^\star$ is the gold answer, and $\mathrm{eval}(\cdot,\cdot)$ denotes containment exact match (CEM),
% returning $1$ if the prediction contains the gold answer and $0$ otherwise.
where $\hat r_{\mathcal{T}} \sim \mathcal{T}(\cdot \mid \mathbf{D}_q,q)$ is the teacher response and $\mathrm{eval}(\cdot,\cdot) \in \{0,1\}$ checks whether the prediction matches the gold answer $r^\star$.
Specifically, we use Gemini 3 Flash to perform this check.

PD operates only on the teacher-correct subset $\mathcal{S}^{+}$, while $\mathcal{S}^{-}$ is reserved for HE.
% PD operates only on the correct slice $\mathcal{S}^{+}$, while $\mathcal{S}^{-}$ is reserved for the second stage.
For each $(q, \mathbf{D}_q) \in \mathcal{S}^{+}$, the teacher reads the original retrieved tokens while the student reads their compressed counterparts $\mathbf{e}_i = \mathcal{C}_\phi(d_i)$, and we minimize the Kullback--Leibler divergence between the two resulting response distributions:
\begin{equation}
    \mathcal{L}_{\text{PD}}(\phi, \theta)
    =
    \mathbb{E}_{(q,\mathbf{D}_q)\sim\mathcal{S}^{+}}
    \Big[
        D_{\mathrm{KL}}\!\big(
        P_{\mathcal{T}}(r \mid \mathbf{D}_q, q)
        \,\big\Vert\,
        P_{\theta}(r \mid \mathbf{e}_1,\ldots,\mathbf{e}_k,q)
        \big)
    \Big].
    \label{eq:pd-loss}
\end{equation}
Notably, the gradient flows through both $\theta$ and, via $\mathbf{e}_i$, through $\phi$.

% \paragraph{Why filter teacher-failed examples?}
% Standard distillation treats the full-context teacher as a reliable target throughout the training distribution.
% DEX-Comp instead uses distillation only where the teacher provides a correct demonstration.
% For teacher-correct examples, imitation provides a useful signal for learning to answer through compressed representations.
% For teacher-failed examples, however, reproducing the teacher's behavior is undesirable.
% We therefore reserve these examples for direct outcome optimization in the second stage.
\paragraph{Why filter teacher-failed examples?}
Teacher-correct responses provide reliable demonstrations for learning through compressed representations, whereas imitating teacher-failed responses directly reinforces undesirable behavior.
We therefore reserve the latter for outcome optimization in Stage~II.
% \paragraph{Rationale.}
% The filtering to $\mathcal{S}^{+}$ reflects a fundamentally different rationale from prior work.
% Existing methods take the uncompressed model as an oracle and try to fully clone its behavior over the entire training distribution.
% In contrast, we treat imitation merely as a fast route to a competent initialization.
% Cloning is neither necessary nor desirable in our setting, since the compressor and decoder operate on representations that are qualitatively different from original tokens, and their optimal computation strategy does not necessarily coincide with that of the teacher.
% We therefore expose the student only to the teacher's reliable demonstrations and leave the search for a more suitable computation strategy for compressed inputs to the second stage.

\subsection{Training Stage II: Hard Exploration}
\label{sec:he}

Pure Distillation provides a reliable initialization but learns only from teacher-correct behavior.
For $\mathcal{S}^{-}$, neither the incorrect teacher responses nor direct fine-tuning on short reference answers provides a desirable imitation target, as the latter can distort the model's response distribution~\citep{ACC-RAG}.
We therefore use reference answers as outcome signals rather than generation targets.
Starting from the PD checkpoint, Hard Exploration (HE) trains on $\mathcal{S}^{-}$ to directly optimize answer correctness.

% Pure Distillation provides a reliable initialization, but it learns only from teacher-correct behavior.
% % \sz{new: }
% For questions in $\mathcal{S}^{-}$, the teacher no longer provides a desirable
% imitation target. Directly fine-tuning on reference answers is also undesirable,
% as short references can impose a response distribution that differs substantially
% from the model's natural answering behavior~\citep{ACC-RAG}.
% However, these references still provide a reliable outcome signal.
% We therefore use them as rewards rather than generation targets, enabling
% optimization for answer correctness without requiring the model to imitate
% either an incorrect teacher response or the reference-answer style.
% Starting from the PD checkpoint, Hard Exploration (HE) trains exclusively on $\mathcal{S}^{-}$ to optimize answer correctness rather than teacher agreement.

Specifically, we adopt Group Relative Policy Optimization (GRPO)~\citep{grpo}.
For each query $q \in \mathcal{S}^{-}$, we draw a group of $G$ candidate responses from the current student policy $\pi_{\phi,\theta}$:
\begin{equation}
    r^{(g)} \sim \pi_{\phi,\theta}\big(\,\cdot \mid \mathbf{e}_1, \ldots, \mathbf{e}_k, q\,\big),
    \qquad
    \mathbf{e}_i = \mathcal{C}_\phi(d_i),
    \qquad
    g = 1, \ldots, G,
\end{equation}
and assign each rollout a binary outcome reward
\begin{equation}
    R^{(g)} = \mathrm{eval}(r^{(g)}, r^\star) \in \{0,1\},
\end{equation}
Specifically, we adopt containment exact match (CEM) as the reward: a rollout receives $1$ if it contains the gold answer and $0$ otherwise.
We cap each rollout at $32$ generated tokens to constrain response length.
Group-relative advantages are then formed by standardizing rewards within each group:
\begin{equation}
    A^{(g)} = \frac{R^{(g)} - \mu_R}{\sigma_R + \epsilon},
    \qquad
    \mu_R = \frac{1}{G}\sum_{g=1}^{G} R^{(g)},
    \qquad
    \sigma_R^2 = \frac{1}{G}\sum_{g=1}^{G}\big(R^{(g)} - \mu_R\big)^2.
\end{equation}

The student is then updated with the clipped policy-gradient surrogate, regularized toward the PD checkpoint $\pi_{\text{ref}}$:
\begin{equation}
\begin{aligned}
    \mathcal{L}_{\text{HE}}(\phi, \theta)
    \;=\;
    -\mathbb{E}_{q\sim\mathcal{S}^{-}}
    \Bigg[
        \frac{1}{G}\sum_{g=1}^{G}
        \frac{1}{|r^{(g)}|}
        \sum_{t=1}^{|r^{(g)}|}
        \min\Big(
            \rho_t^{(g)} A^{(g)},
            \mathrm{clip}\big(\rho_t^{(g)}, 1-\varepsilon, 1+\varepsilon\big) A^{(g)}
        \Big)
    \Bigg]
    \\
    + \beta\, D_{\mathrm{KL}}\!\left[\pi_{\phi,\theta}\,\Vert\,\pi_{\text{ref}}\right],
\end{aligned}
\label{eq:he-loss}
\end{equation}
where
\begin{equation}
    \rho_t^{(g)}
    =
    \frac{
        \pi_{\phi,\theta}(r_t^{(g)} \mid r_{<t}^{(g)}, q, \mathbf{e}_{1:k})
    }{
        \pi_{\text{old}}(r_t^{(g)} \mid r_{<t}^{(g)}, q, \mathbf{e}_{1:k})
    },
\end{equation}
$\varepsilon$ is the clipping range, and $\beta$ controls the KL regularization.

Restricting HE to $\mathcal{S}^{-}$ is deliberate:
these are precisely the examples for which teacher imitation provides an unreliable learning signal.
Rather than reproducing the teacher's failed behavior, HE directly rewards successful responses generated from compressed inputs.
Together, PD and HE assign imitation and outcome optimization to the subsets where each learning signal is most appropriate.

\section{Experimental Setup}
\label{sec:exp_setup}

\subsection{Datasets and Evaluation}
We train and evaluate on five open-domain QA datasets:
Natural Questions~\citep{nq} and TriviaQA~\citep{trivalqa} for factoid QA,
HotpotQA~\citep{hotpotqa} for multi-hop QA,
ASQA~\citep{asqa} for ambiguous long-form QA, and
PopQA~\citep{popqa} for long-tail QA.
We use the Wikipedia-KILT corpus preprocessed by~\citet{cocom} for retrieval.

We evaluate retrieval depths from top-$5$ to top-$30$ using two metrics:
(i) \textbf{Containment Exact Match (CEM)}, which checks whether a ground-truth answer appears in the prediction~\citep{xrag,pisco};
and (ii) \textbf{LLM-as-judge}, where Gemini 3 Flash\footnote{\url{https://ai.google.dev/gemini-api/docs/models/gemini-3-flash-preview}} judges prediction correctness given the ground-truth answers.
The judging prompt is provided in Appendix~\ref{sec:llm_judge_prompt}.

\subsection{Baseline Methods}
\label{sec:baseline_methods}
% We compare against uncompressed RAG and query-independent compression methods.
% The latter include LLMLingua-2~\citep{llmlingua2} for hard compression and
% xRAG~\citep{xrag}, ICAE~\citep{icae}, COCOM~\citep{cocom},
% SAC~\citep{autoencoding}, RLComp~\citep{ting2026bridging}, and
% PISCO~\citep{pisco} for soft compression.
% All methods use Mistral-7B~\citep{mistral} as the decoder backbone.
We compare DEX-Comp against two groups of baselines.
(i) \textbf{Uncompressed RAG.}
This baseline feeds the full retrieved context into the language model without compression, serving as the reference for measuring the compression gap.
(ii) \textbf{Query-independent compression methods.}
Like DEX-Comp, these methods compress documents without access to the query, allowing document representations to be computed independently of individual queries.
We include LLMLingua-2~\citep{llmlingua2} as a hard-compression baseline, alongside the soft-compression methods xRAG~\citep{xrag}, ICAE~\citep{icae}, COCOM~\citep{cocom}, SAC~\citep{autoencoding}, RLComp~\citep{ting2026bridging}, and PISCO~\citep{pisco}.
To improve comparability, all methods use Mistral-7B~\citep{mistral} as the decoder backbone.

\subsection{Implementation Details}
\label{sec:impl}
We use BERGEN~\citep{bergen} for preprocessing and retrieval, with
Splade-v3~\citep{splade} as the retriever and DeBERTa-v3~\citep{DeBERTa} as the reranker.
All methods use Mistral-7B-Instruct~\citep{mistral} as the backbone LLM with temperature $0$.
For DEX-Comp, the compressor and decoder share the backbone and use separate LoRA~\citep{lora} adapters.

We combine the training splits of all five datasets into a unified training pool and generate a full-context teacher response for each question.
Using Gemini 3 Flash, we partition the pool into teacher-correct $\mathcal{S}^{+}$ and teacher-failed $\mathcal{S}^{-}$ examples as defined in Eq.~\ref{eq:partition}.
PD trains on $\mathcal{S}^{+}$, and HE initializes from PD and trains on $\mathcal{S}^{-}$.
We train separate variants across retrieval depths and compression rates; unless otherwise specified, experiments use $16\times$ compression with matched training and inference depths.
All experiments run on 8 RTX PRO 6000 Blackwell GPUs using PyTorch and Transformers.
\section{Experimental Results}
\label{sec:exp_results}

\subsection{Main Results}
\label{sec:main_results}
\begin{table}[t]
    \centering
% Previous wording retained for review:
%     \caption{Results at different retrieval depths (top-5 to top-30). For each retrieval depth, the best result is in bold, and the second-best is underlined. All methods perform inference with Mistral-7B. Models marked with $^\dagger$ are retrained at the corresponding top-$k$ since checkpoints for the required configurations are unavailable.}
    \caption{Results with Mistral-7B at top-$5/15/30$. Best and second-best scores at each depth are bold and underlined. $^\dagger$: retrained at the corresponding top-$k$ because matching checkpoints are unavailable.}
    \label{tab:results-comparison}
    \small
    \setlength{\tabcolsep}{3pt}
    \resizebox{\textwidth}{!}{
    \begin{tabular}{lccccccccccccc}
    \toprule
    \multirow{2}{*}{\textbf{Method}} & \multirow{2}{*}{\textbf{Comp. Rate}} & \multicolumn{2}{c}{\textbf{NQ}} & \multicolumn{2}{c}{\textbf{TriviaQA}} & \multicolumn{2}{c}{\textbf{HotpotQA}} & \multicolumn{2}{c}{\textbf{ASQA}} & \multicolumn{2}{c}{\textbf{PopQA}} & \multicolumn{2}{c}{\textbf{Avg.}} \\
    \cmidrule(lr){3-4} \cmidrule(lr){5-6} \cmidrule(lr){7-8} \cmidrule(lr){9-10} \cmidrule(lr){11-12} \cmidrule(lr){13-14}
    & & CEM & LLM & CEM & LLM & CEM & LLM & CEM & LLM & CEM & LLM & CEM & LLM \\
\midrule
    \multicolumn{14}{c}{\textbf{Top-5}} \\
    \midrule
    Full context & - & 58.65 & \underline{73.67} & 90.18 & \underline{89.04} & \underline{46.45} & \underline{55.57} & 68.99 & 77.00 & \underline{58.53} & \underline{57.36} & \underline{64.56} & \underline{70.53} \\
    LLMLingua-2 & $4\times$ & 49.74 & 63.59 & 84.41 & 83.19 & 36.43 & 45.50 & 59.49 & 65.30 & 39.07 & 39.70 & 53.83 & 59.46 \\
    xRAG & $128\times$ & 37.36 & 51.46 & 78.06 & 78.54 & 27.66 & 35.27 & 43.25 & 50.63 & 28.15 & 29.12 & 42.90 & 49.00 \\
    ICAE & $4\times$ & 41.59 & 56.33 & 77.78 & 79.27 & 28.75 & 40.48 & 47.68 & 60.65 & 38.28 & 40.24 & 46.82 & 55.39 \\
    COCOM & $16\times$ & 30.35 & 45.05 & 67.99 & 71.22 & 23.25 & 34.71 & 37.87 & 48.31 & 19.74 & 22.37 & 35.84 & 44.33 \\
    SAC$^\dagger$ & $16\times$ & 51.02 & 65.85 & 86.43 & 85.67 & 41.34 & 50.38 & 59.52 & 66.17 & 41.49 & 42.05 & 55.96 & 62.02 \\
    RLComp$^\dagger$ & $16\times$ & 56.92 & 71.63 & 88.61 & 87.15 & 44.33 & 53.04 & 57.07 & 64.29 & 38.20 & 38.56 & 57.03 & 62.93 \\
    PISCO & $16\times$ & 56.61 & 70.81 & 88.01 & 87.06 & 43.88 & 51.80 & 66.24 & 74.26 & 55.95 & 54.78 & 62.14 & 67.74 \\
    DEX-Comp & $128\times$ & \underline{59.29} & 73.46 & \underline{91.32} & 87.92 & 44.52 & 53.82 & \underline{69.51} & \underline{77.64} & 50.80 & 49.63 & 63.09 & 68.49 \\
    DEX-Comp & $16\times$ & \textbf{62.71} & \textbf{76.52} & \textbf{92.72} & \textbf{89.97} & \textbf{51.04} & \textbf{57.55} & \textbf{75.21} & \textbf{79.96} & \textbf{60.22} & \textbf{58.70} & \textbf{68.38} & \textbf{72.54} \\
\midrule
    \multicolumn{14}{c}{\textbf{Top-15}} \\
    \midrule
    Full context & - & \underline{58.34} & \underline{73.49} & \underline{89.65} & \underline{89.26} & \underline{45.79} & \underline{54.91} & \underline{68.67} & \underline{78.16} & \underline{57.24} & \underline{56.88} & \underline{63.94} & \underline{70.54} \\
    LLMLingua-2 & $4\times$ & 50.30 & 65.00 & 84.71 & 82.70 & 35.66 & 44.95 & 60.02 & 66.35 & 38.75 & 38.18 & 53.89 & 59.43 \\
    PISCO$^\dagger$ & $16\times$ & 55.90 & 70.99 & 87.13 & 86.14 & 41.20 & 49.98 & 65.61 & 74.58 & 51.42 & 50.20 & 60.25 & 66.38 \\
    DEX-Comp & $16\times$ & \textbf{63.27} & \textbf{75.61} & \textbf{91.84} & \textbf{90.38} & \textbf{50.36} & \textbf{59.30} & \textbf{73.84} & \textbf{80.27} & \textbf{59.87} & \textbf{58.84} & \textbf{67.83} & \textbf{72.88} \\
\midrule
    \multicolumn{14}{c}{\textbf{Top-30}} \\
    \midrule
    Full context & - & \underline{57.42} & \underline{72.75} & \underline{88.98} & \underline{88.20} & \underline{45.45} & \underline{54.64} & \underline{67.41} & \underline{76.79} & \underline{55.30} & \underline{55.04} & \underline{62.91} & \underline{69.49} \\
    LLMLingua-2 & $4\times$ & 49.35 & 63.98 & 83.42 & 81.50 & 34.64 & 43.89 & 60.55 & 67.41 & 37.74 & 31.28 & 53.14 & 57.61 \\
    PISCO$^\dagger$ & $16\times$ & 51.99 & 67.57 & 86.69 & 86.59 & 40.46 & 50.20 & 64.24 & 72.68 & 46.12 & 46.89 & 57.90 & 64.79 \\
    DEX-Comp & $16\times$ & \textbf{60.31} & \textbf{75.57} & \textbf{91.62} & \textbf{90.12} & \textbf{48.16} & \textbf{58.48} & \textbf{73.84} & \textbf{80.38} & \textbf{56.23} & \textbf{55.53} & \textbf{66.03} & \textbf{72.02} \\
    \bottomrule
    \end{tabular}}
\end{table}

% Following prior work, we first evaluate DEX-Comp with training and inference using the same top-$k$.
% As shown in Table~\ref{tab:results-comparison}, at $16\times$ compression DEX-Comp surpasses the uncompressed RAG baseline on every dataset and retrieval depth, with statistically significant gains ($p<0.05$, paired permutation test) and average CEM\,/\,LLM improvements above two points ($+3.82/+2.01$, $+3.89/+2.34$, $+3.12/+2.53$ at $k\!=\!5/15/30$).
With matched training and inference depths (Table~\ref{tab:results-comparison}), DEX-Comp at $16\times$ compression surpasses uncompressed RAG on every dataset and depth, with average CEM\,/\,LLM gains above two points ($p<0.05$, paired permutation test).
To our knowledge, this is the first query-independent soft compression method shown to surpass the uncompressed RAG model across multiple realistic retrieval depths, establishing a new state of the art among comparable compression methods.

\subsection{Inference Efficiency}
\label{sec:exp_efficiency}
\begin{table}[t]
    \centering
% Previous wording retained for review:
%     \caption{Inference efficiency comparison between RAG and DEX-Comp across different top-$k$ retrieval settings. We report Time-To-First-Token (TTFT), computational cost (GFLOPs), and peak GPU memory (MB).}
    \caption{Inference efficiency: TTFT, computation (GFLOPs), and peak GPU memory (MB).}
    \label{tab:efficiency}
% Previous wording retained for review:
%     \vskip 0.15in
    \begin{small}
    \resizebox{\textwidth}{!}{
    \begin{tabular}{lccccccccc}
    \toprule
    \multirow{2}{*}{\textbf{Top-$k$}} & \multicolumn{3}{c}{\textbf{TTFT (ms)}} & \multicolumn{3}{c}{\textbf{GFLOPs}} & \multicolumn{3}{c}{\textbf{Memory (MB)}} \\
    \cmidrule(lr){2-4} \cmidrule(lr){5-7} \cmidrule(lr){8-10}
    & RAG & DEX-Comp & Speedup & RAG & DEX-Comp & Reduction & RAG & DEX-Comp & Reduction \\
    \midrule
    Top-5  & 373   & 84  & \textbf{4.42$\times$}  & 82{,}850  & 12{,}423 & \textbf{6.67$\times$}  & 16{,}100 & 14{,}242 & \textbf{1.13$\times$} \\
    Top-15 & 1{,}626 & 131 & \textbf{12.33$\times$} & 245{,}650 & 22{,}654 & \textbf{10.84$\times$} & 26{,}887 & 14{,}457 & \textbf{1.86$\times$} \\
    Top-30 & 4{,}782 & 201 & \textbf{23.73$\times$} & 518{,}767 & 38{,}129 & \textbf{13.61$\times$} & 59{,}328 & 14{,}766 & \textbf{4.02$\times$} \\
    \bottomrule
    \end{tabular}
    }
    \end{small}
\end{table}

% We measure inference efficiency on a single RTX PRO 6000 Blackwell GPU with batch size $8$, reporting Time-To-First-Token (TTFT), computational cost (GFLOPs), and peak GPU memory in Table~\ref{tab:efficiency}.
% DEX-Comp improves all three measures relative to full-context RAG, with larger gains at greater retrieval depths.
% At top-$30$, it reduces peak GPU memory by $4.02\times$, computational cost by $13.61\times$, and TTFT by $23.73\times$, demonstrating substantial efficiency improvement.
On a single RTX PRO 6000 Blackwell GPU with batch size $8$, DEX-Comp improves Time-To-First-Token (TTFT), GFLOPs, and peak memory over full-context RAG, with larger gains at greater retrieval depths (Table~\ref{tab:efficiency}).
% \paragraph{Offline cost.}
% DEX-Comp requires a one-time offline compression pass over the corpus and storage of compressed embeddings.
% In our setup, corpus compression is performed once and amortized across all queries.
% The resulting embedding remains fixed unless the corpus is updated, and for dynamic corpora, incremental recompression can be applied only to updated documents.
Offline compression and embedding storage incur a one-time corpus-level cost amortized across queries; only updated documents require recompression.
% Notebly, DEX-Comp requires one-time offline corpus compression and embedding storage.
% The cost is amortized across queries, with recompression limited to updated documents.

\section{Analysis}
\label{sec:analysis}

\subsection{Ablation Study}
\label{sec:ablation}

\begin{table}[t]
    \centering
% Previous wording retained for review:
%     \caption{Ablation at top-30. Dataset scores and Acc report LLM-judged accuracy (\%). Success, failure, and mixed denote RL subsets containing teacher-correct questions, teacher-incorrect questions, or both.}
    \caption{Top-$30$ ablation. Dataset scores and Acc are LLM-judged accuracy (\%). RL subsets contain teacher-correct (Success), teacher-failed (Failure), or both (Mixed) questions.}
    \label{tab:ablation}
% Previous wording retained for review:
%     \vskip 0.10in
    \begin{small}
    \resizebox{\textwidth}{!}{
    \begin{tabular}{llcccccccc}
    \toprule
    \multirow{2}{*}{\textbf{Initialization}} & \multirow{2}{*}{\textbf{RL subset}} & \multirow{2}{*}{\textbf{NQ}} & \multirow{2}{*}{\textbf{TriviaQA}} & \multirow{2}{*}{\textbf{HotpotQA}} & \multirow{2}{*}{\textbf{ASQA}} & \multirow{2}{*}{\textbf{PopQA}} & \multicolumn{3}{c}{\textbf{Average}} \\
    \cmidrule(lr){8-10}
    & & & & & & & Acc & Resilience & Boost \\
    \midrule
    Standard distill. & None & 67.57 & 86.59 & 50.20 & 72.68 & 46.89 & 64.79 & 85.25 & 45.96 \\
    PD & None (w/o HE) & 70.96 & 88.74 & 55.32 & 74.79 & 48.95 & 67.75 & 87.63 & 49.56 \\
    \midrule
    Standard distill. & Failure (HE) & 69.46 & 86.83 & 51.85 & 74.22 & 49.67 & 66.41 & 87.01 & 48.16 \\
    PD & Success & 74.41 & 90.11 & 57.30 & 77.43 & 53.74 & 70.60 & 89.58 & 53.31 \\
    PD & Mixed & 75.15 & 90.10 & 58.13 & 78.90 & 55.07 & 71.47 & 90.13 & 54.49 \\
    \textbf{PD (DEX-Comp)} & \textbf{Failure (HE)} & \textbf{75.57} & \textbf{90.12} & \textbf{58.48} & \textbf{80.38} & \textbf{55.53} & \textbf{72.02} & \textbf{90.78} & \textbf{55.00} \\
    \bottomrule
    \end{tabular}
    }
    \end{small}
\end{table}

Table~\ref{tab:ablation} validates both stages of DEX-Comp and the use of teacher correctness to allocate different learning signals.
PD improves average LLM-judged accuracy from $64.79$ to $67.75$ before RL, confirming the benefit of excluding teacher-failed examples from distillation.
Applying HE to teacher-failed examples further improves accuracy to $72.02$ after PD, compared with $66.41$ when initialized from standard distillation, suggesting that PD provides a stronger initialization for exploration.
Importantly, starting from the same PD checkpoint, exploration on teacher-failed examples achieves the highest accuracy ($72.02$), outperforming exploration on teacher-correct examples ($70.60$) or their mixture ($71.47$).
Together, these results support the central design of DEX-Comp: teacher correctness is not merely a filtering criterion for distillation, but provides a useful signal for assigning imitation and outcome optimization to different training examples.
Both stages also improve Resilience and Boost, especially Boost (Section~\ref{sec:analysis_beyond_overall_score}), indicating more effective retrieval use; Figure~\ref{fig:distillation_limits}(b,c) provides complementary diagnostics on teacher-failed questions.

\subsection{Context Compression for Efficient RL}
\label{sec:training_cost}

\begin{table}[t]
    \centering
% Previous wording retained for review:
%     \caption{Top-$5$ LLM-judged accuracy (\%) and training cost on eight GPUs. GPU-hours denotes the total training gpu time. Max. RL batch denotes the largest per-GPU batch that fits in memory.}
    \caption{Top-$5$ LLM-judged accuracy (\%) and training cost on eight GPUs. Max. RL batch is the largest per-GPU batch fitting in memory; GPU-hours count total GPU training time.}
    \label{tab:training_cost}
    \small
    \setlength{\tabcolsep}{4pt}
    \resizebox{\linewidth}{!}{%
    \begin{tabular}{@{}lcccccccc@{}}
    \toprule
    Configuration & NQ & TriviaQA & HotpotQA & ASQA & PopQA & Avg. & GPU-hours & Max. RL batch \\
    \midrule
    Full context & $73.67$ & $89.04$ & $55.57$ & $77.00$ & $57.36$ & $70.53$ & -- & -- \\
    Full context + RL & $80.23$ & $92.36$ & $65.00$  & $82.28$ & $64.02$ & $76.78$ & $38$ & $16$ \\
    PD ($16\times$) & $71.04$ & $88.25$  & $54.72$ & $74.18$  & $51.41$ & $67.92$ & $1.5$ & -- \\
    PD + HE ($16\times$) & $76.52$ & $89.97$ & $57.55$ & $79.96$ & $58.70$ & $72.54$ & $13$ & $64$ \\
    \bottomrule
    \end{tabular}%
    }
\end{table}

% Previous wording retained for review:
% We compare DEX-Comp with the untuned full-context model and a full-context variant trained with the same RL data and procedure as HE (Table~\ref{tab:training_cost}).
% Both RL runs use eight GPUs and the largest per-GPU batch size that fits in memory.
% At top-$5$, compressed inputs support a $4\times$ larger per-GPU RL batch ($64$ vs.\ $16$).
% Including PD, DEX-Comp requires $13$ GPU-hours, $65.8\%$ less than full-context RL ($38$ GPU-hours).
We compare DEX-Comp with untuned full-context RAG and full-context RL using matched RL data and procedures on eight GPUs (Table~\ref{tab:training_cost}).
At top-$5$, compression supports a $4\times$ larger maximum per-GPU batch ($64$ vs.\ $16$) and uses $65.8\%$ fewer GPU-hours ($13$, including PD, vs.\ $38$).
It improves average accuracy over the untuned full-context model from $70.53$ to $72.54$, while full-context RL reaches $76.78$.
% Previous wording retained for review:
% At top-$30$, where full-context RL exceeds our memory budget, compressing before RL remains feasible and improves accuracy over untuned full-context RAG ($72.02\%$ vs.\ $69.49\%$).
% % Previous wording retained for review:
% Thus, compressing before RL both reduces training cost and extends the feasible context length under the same hardware constraints.
Thus, compression reduces RL cost and enables training at top-$30$, where full-context RL exceeds our memory budget, while improving accuracy over untuned RAG ($72.02\%$ vs.\ $69.49\%$).

\subsection{RAG Behavior Analysis}
\label{sec:analysis_beyond_overall_score}

\begin{figure}[!htb]
    \centering
    \includegraphics[width=0.9\textwidth]{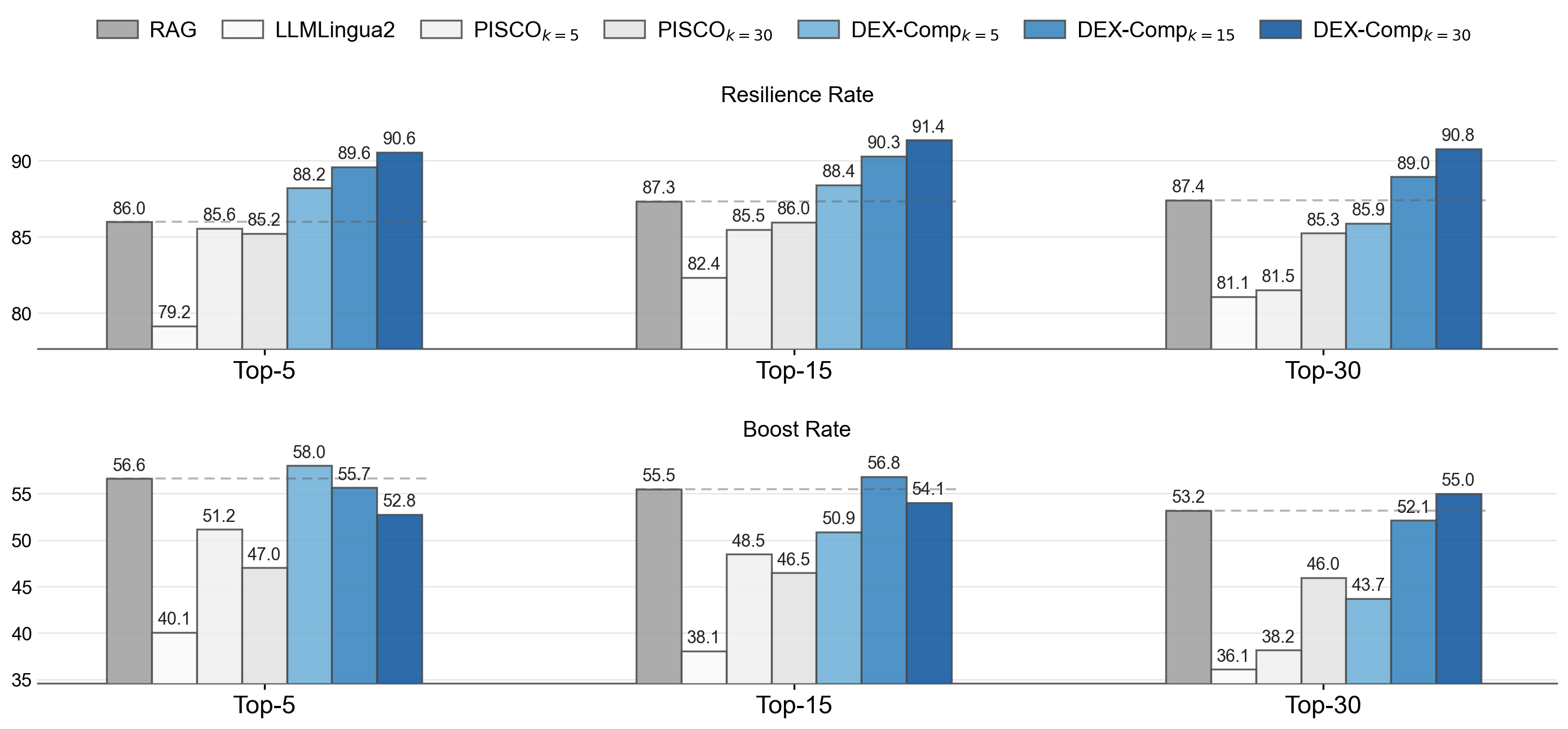}
% Previous wording retained for review:
%     \caption{Resilience Rate (top) and Boost Rate (bottom) at retrieval depths $k\in\{5,15,30\}$. The subscript $k$ indicates the top-$k$ used during training.}
    \caption{Resilience Rate (top) and Boost Rate (bottom) across retrieval depths. Subscripts denote the training top-$k$.}
    \label{fig:rag_metric}
\end{figure}

% Previous wording retained for review:
% Beyond overall accuracy, we analyze performance using two RAG metrics~\cite{xrag}.
% The \emph{Resilience Rate} measures correctness with and without retrieval, reflecting robustness to noisy context.
% The \emph{Boost Rate} measures cases where retrieval corrects an initially incorrect answer, reflecting effective use of retrieved information.
% Figure~\ref{fig:rag_metric} shows that DEX-Comp surpasses the uncompressed RAG reference on both metrics when training and inference depths match, indicating stronger robustness and information utilization.
Resilience Rate measures correctness with and without retrieval, while Boost Rate measures retrieval-based corrections to initially incorrect answers~\citep{xrag}.
DEX-Comp surpasses uncompressed RAG on both metrics at matched training and inference depths (Figure~\ref{fig:rag_metric}).
Gains are larger for Resilience Rate, suggesting that compression acts as an implicit denoising mechanism.

\subsection{Sensitivity to Compression and Retrieval Settings}
\label{sec:training_parameters}

\begin{figure}[!htb]
    \centering
    \includegraphics[width=0.9\textwidth]{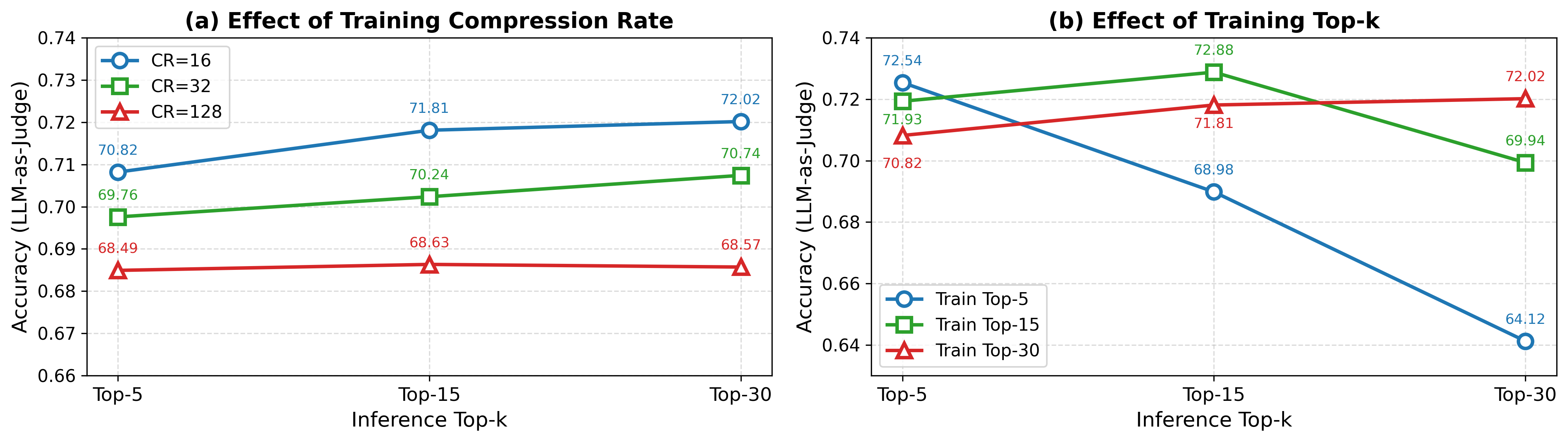}
% Previous wording retained for review:
%     \caption{Effect of (a) the compression rate (CR) and (b) the retrieval depth during training. Each panel varies one parameter while keeping the other fixed.}
    \caption{Effects of (a) compression rate and (b) training depth, with the other fixed.}
    \label{fig:cr_topk}
\end{figure}

% Previous wording retained for review:
% Figure~\ref{fig:cr_topk} studies key training parameters.
% Accuracy improves as the compression rate decreases, with CR\,=\,16 performing best across the tested retrieval depths; higher rates offer shorter inputs at a cost in accuracy.
% For retrieval depth, performance is highest when training and inference match.
% Training at larger $k$ transfers more reliably to smaller retrieval depths than the reverse, suggesting that larger training depths are preferable when deployment depth varies.
Lower compression rates improve accuracy, with CR\,=\,16 best across tested retrieval depths (Figure~\ref{fig:cr_topk}).
Matching training and inference depths performs best; training at larger $k$ transfers more reliably to smaller depths than the reverse, favoring larger $k$ when deployment depth varies.

\subsection{Generalization across Backbones and Datasets}
\label{sec:generalization}

\begin{figure}[!htbp]
    \centering
    \includegraphics[width=0.9\textwidth]{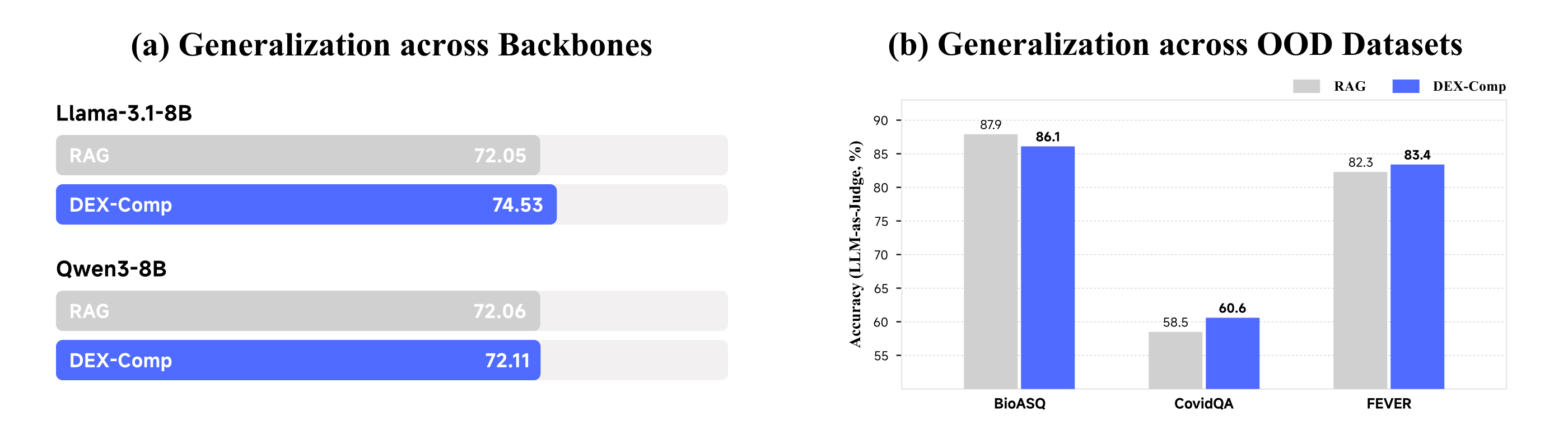}
% Previous wording retained for review:
%     \caption{Generalization of DEX-Comp (a) across backbone model families and (b) across out-of-distribution datasets.}
    \caption{Generalization across (a) backbone families and (b) out-of-distribution datasets.}
    \label{fig:generalization}
\end{figure}

% Previous wording retained for review:
% We evaluate generalization across both model architectures and data distributions.
Figure~\ref{fig:generalization}(a) shows an improvement over the corresponding uncompressed RAG reference with Llama-3.1-8B and Qwen3-8B.
For out-of-distribution evaluation, DEX-Comp performs better on CovidQA~\citep{COVID-QA} and FEVER~\citep{Fever}, but lower on BioASQ~\citep{bioasq} (Figure~\ref{fig:generalization}(b)).
% Previous wording retained for review:
% These results show that our training recipe generalizes across backbones and datasets.

\subsection{Analysis of Compression Embeddings}
\label{sec:compression_embeddings}

\begin{figure}[!htbp]
    \centering
    \includegraphics[width=0.82\textwidth,height=0.20\textheight,keepaspectratio]{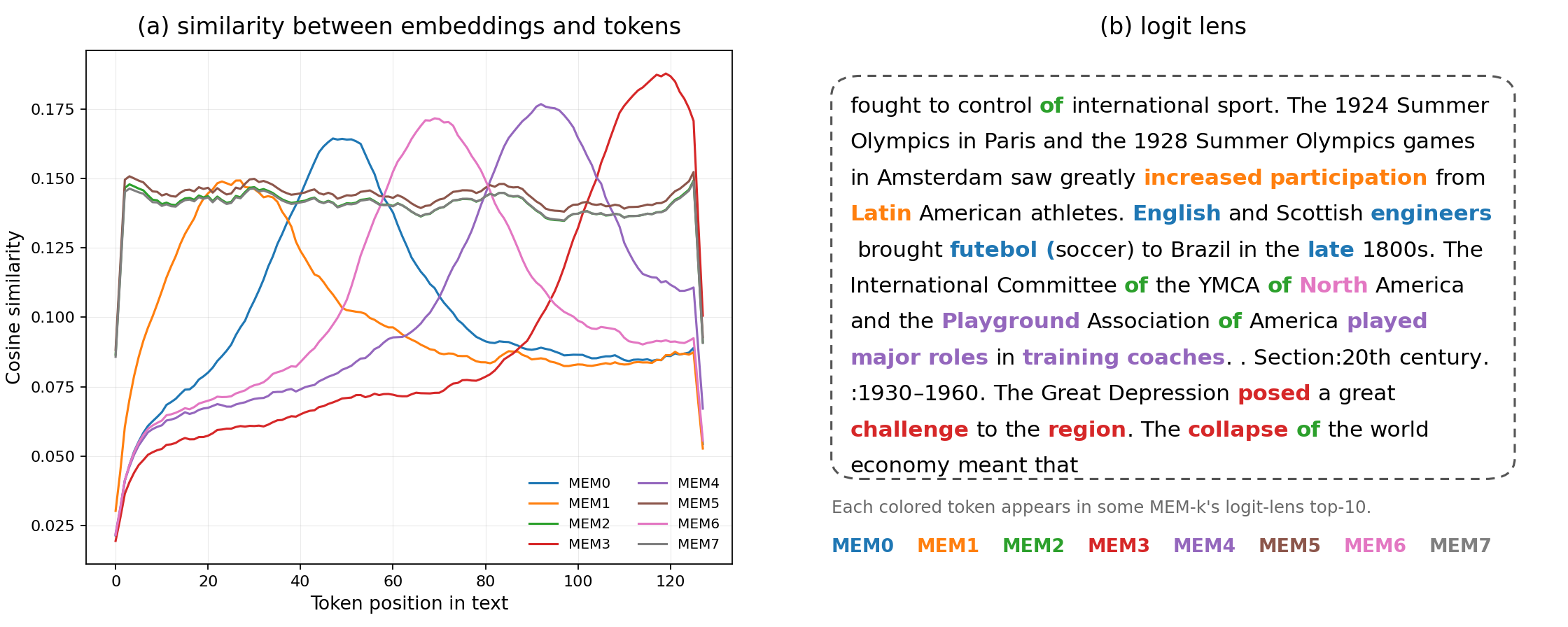}
% Previous wording retained for review:
%     \caption{Probing DEX-Comp's compression embeddings. (a) Cosine similarity between embeddings and tokens. (b) Logit lens projections highlighting top tokens associated with each embedding.}
    \caption{Compression embeddings: (a) token cosine similarity; (b) top logit-lens tokens.}
    \label{fig:embeddings}
\end{figure}

We analyze learned embeddings using cosine similarity and logit lens projections~\citep{logitlens}.
Embeddings exhibit spatial specialization, with some focusing on specific document regions and others capturing global information, suggesting complementary roles.
Compared with PISCO trained by standard distillation, DEX-Comp exhibits markedly different embedding representations.
The comparison is detailed in Appendix~\ref{sec:pisco_embedding_comparison}.

% Limitations is not a required section under the ICLR 2027 Author Guidelines.
% Removed from the rendered paper at the author's request; original text retained below.
% \section{Limitations}
% \label{sec:limitations}

% DEX-Comp assumes a query-independent, offline compression setting, which requires precomputing and storing compressed embeddings for the corpus.
% While this cost is amortized across queries, it may become significant for frequently updated or very large-scale corpora.

% Our comparison uses an uncompressed RAG baseline without task-specific fine-tuning or reinforcement learning, while DEX-Comp is trained with distillation and RL signals derived from gold answers. As a result, the observed gains do not isolate the effect of compression alone. A more controlled comparison would apply similar training to the uncompressed baseline, which we leave for future work. 
% Given the inherent difficulty of disentangling the effects of compression from task-specific RL, our current setup does not strictly isolate these gains.
% A more controlled comparison is left for future work.
% The full-context RL comparison provides a task-trained reference, but the gains over untuned RAG do not isolate the effect of compression from task-specific learning.
% Finally, our evaluation focuses on QA benchmarks and answer correctness metrics. While we observe strong empirical performance, future work should investigate faithfulness, grounding, and robustness under adversarial or long-form generation settings.

\section{Conclusion}
\label{sec:conclusion}

We propose DEX-Comp, a two-stage training recipe for soft context compression that combines Pure Distillation with Hard Exploration.
The recipe improves on the evaluated compression baselines without separate pretraining.
At top-30 retrieval, DEX-Comp compresses the retrieved context by $16\times$ and accelerates inference by over $20\times$, while achieving performance comparable to or exceeding uncompressed RAG across five benchmarks.
Further analyses confirm the contribution of each stage and demonstrate that DEX-Comp generalizes across backbone models and data distributions.

We believe this work provides a step toward understanding and improving soft context compression, and toward narrowing the gap between compressed and uncompressed RAG systems.

% ICLR 2027 requires an AI use statement; it does not count toward the page limit.
% Adapted from https://iclr.cc/Conferences/2027/AIPolicyForAuthors.
\clearpage
\subsection*{AI use statement}
Our experimental use of LLMs includes teacher-response generation and Gemini 3 Flash judging for training-set partitioning and evaluation, as described in Sections 3–4 and Appendix A. 
Beyond these experimental uses, generative AI assistance was limited to sentence-level language editing, initial figure drafts, and experimental code. 
The authors take full responsibility for the final content of this paper, including its text, figures, code, and reported results.

\bibliography{iclr2027_conference}

@inproceedings{xrag,
  author       = {Xin Cheng and
                  Xun Wang and
                  Xingxing Zhang and
                  Tao Ge and
                  Si{-}Qing Chen and
                  Furu Wei and
                  Huishuai Zhang and
                  Dongyan Zhao},
  editor       = {Amir Globersons and
                  Lester Mackey and
                  Danielle Belgrave and
                  Angela Fan and
                  Ulrich Paquet and
                  Jakub M. Tomczak and
                  Cheng Zhang},
  title        = {{xRAG}: Extreme Context Compression for Retrieval-augmented Generation
                  with One Token},
  booktitle    = {Advances in Neural Information Processing Systems 38: Annual Conference
                  on Neural Information Processing Systems 2024, NeurIPS 2024, Vancouver,
                  BC, Canada, December 10 - 15, 2024},
  year         = {2024},
  url          = {http://papers.nips.cc/paper\_files/paper/2024/hash/c5cf13bfd3762821ef7607e63ee90075-Abstract-Conference.html},
  bibsource    = {dblp computer science bibliography, https://dblp.org}
}

@inproceedings{DBLP:conf/emnlp/KarpukhinOMLWEC20,
  author       = {Vladimir Karpukhin and
                  Barlas Oguz and
                  Sewon Min and
                  Patrick Lewis and
                  Ledell Wu and
                  Sergey Edunov and
                  Danqi Chen and
                  Wen{-}tau Yih},
  editor       = {Bonnie Webber and
                  Trevor Cohn and
                  Yulan He and
                  Yang Liu},
  title        = {Dense Passage Retrieval for Open-Domain Question Answering},
  booktitle    = {Proceedings of the 2020 Conference on Empirical Methods in Natural
                  Language Processing, {EMNLP} 2020, Online, November 16-20, 2020},
  pages        = {6769--6781},
  publisher    = {Association for Computational Linguistics},
  year         = {2020},
  url          = {https://doi.org/10.18653/v1/2020.emnlp-main.550},
  doi          = {10.18653/V1/2020.EMNLP-MAIN.550},
  bibsource    = {dblp computer science bibliography, https://dblp.org}
}

@inproceedings{Palu,
  author       = {Chi{-}Chih Chang and
                  Wei{-}Cheng Lin and
                  Chien{-}Yu Lin and
                  Chong{-}Yan Chen and
                  Yu{-}Fang Hu and
                  Pei{-}Shuo Wang and
                  Ning{-}Chi Huang and
                  Luis Ceze and
                  Mohamed S. Abdelfattah and
                  Kai{-}Chiang Wu},
  title        = {Palu: {KV-Cache} Compression with Low-Rank Projection},
  booktitle    = {The Thirteenth International Conference on Learning Representations,
                  {ICLR} 2025, Singapore, April 24-28, 2025},
  publisher    = {OpenReview.net},
  year         = {2025},
  url          = {https://openreview.net/forum?id=LWMS4pk2vK},
  bibsource    = {dblp computer science bibliography, https://dblp.org}
}

@inproceedings{SnapKV,
  author       = {Yuhong Li and
                  Yingbing Huang and
                  Bowen Yang and
                  Bharat Venkitesh and
                  Acyr Locatelli and
                  Hanchen Ye and
                  Tianle Cai and
                  Patrick Lewis and
                  Deming Chen},
  editor       = {Amir Globersons and
                  Lester Mackey and
                  Danielle Belgrave and
                  Angela Fan and
                  Ulrich Paquet and
                  Jakub M. Tomczak and
                  Cheng Zhang},
  title        = {{SnapKV}: {LLM} Knows What You are Looking for Before Generation},
  booktitle    = {Advances in Neural Information Processing Systems 38: Annual Conference
                  on Neural Information Processing Systems 2024, NeurIPS 2024, Vancouver,
                  BC, Canada, December 10 - 15, 2024},
  year         = {2024},
  url          = {http://papers.nips.cc/paper\_files/paper/2024/hash/28ab418242603e0f7323e54185d19bde-Abstract-Conference.html},
  bibsource    = {dblp computer science bibliography, https://dblp.org}
}

@inproceedings{OmniKV,
  author       = {Jitai Hao and
                  Yuke Zhu and
                  Tian Wang and
                  Jun Yu and
                  Xin Xin and
                  Bo Zheng and
                  Zhaochun Ren and
                  Sheng Guo},
  title        = {{OmniKV}: Dynamic Context Selection for Efficient Long-Context {LLMs}},
  booktitle    = {The Thirteenth International Conference on Learning Representations,
                  {ICLR} 2025, Singapore, April 24-28, 2025},
  publisher    = {OpenReview.net},
  year         = {2025},
  url          = {https://openreview.net/forum?id=ulCAPXYXfa},
  bibsource    = {dblp computer science bibliography, https://dblp.org}
}

@article{deltakv,
  author       = {Jitai Hao and
                  Qiang Huang and
                  Yaowei Wang and
                  Min Zhang and
                  Jun Yu},
  title        = {{DeltaKV}: Residual-Based {KV} Cache Compression via Long-Range Similarity},
  journal      = {CoRR},
  volume       = {abs/2602.08005},
  year         = {2026},
  url          = {https://doi.org/10.48550/arXiv.2602.08005},
  doi          = {10.48550/ARXIV.2602.08005},
  eprinttype   = {arXiv},
  eprint       = {2602.08005},
  bibsource    = {dblp computer science bibliography, https://dblp.org}
}

@inproceedings{longllmlingua,
  title={Longllmlingua: Accelerating and enhancing llms in long context scenarios via prompt compression},
  author={Jiang, Huiqiang and Wu, Qianhui and Luo, Xufang and Li, Dongsheng and Lin, Chin-Yew and Yang, Yuqing and Qiu, Lili},
  booktitle={Proceedings of the 62nd Annual Meeting of the Association for Computational Linguistics (Volume 1: Long Papers)},
  pages={1658--1677},
  year={2024}
}

@inproceedings{tang2026gmsa,
  title={Gmsa: Enhancing context compression via group merging and layer semantic alignment},
  author={Tang, Jiwei and Zhang, Zhicheng and Wu, Shunlong and Ye, Jingheng and Bai, Lichen and Wang, Zitai and Lu, Tingwei and Hai, Lin and Zhao, Yiming and Zheng, Hai-Tao and others},
  booktitle={Proceedings of the 64th Annual Meeting of the Association for Computational Linguistics (Volume 1: Long Papers)},
  pages={28690--28704},
  year={2026}
}

@inproceedings{pisco,
  author       = {Maxime Louis and
                  Herv{\'{e}} D{\'{e}}jean and
                  St{\'{e}}phane Clinchant},
  editor       = {Wanxiang Che and
                  Joyce Nabende and
                  Ekaterina Shutova and
                  Mohammad Taher Pilehvar},
  title        = {{PISCO:} Pretty Simple Compression for Retrieval-Augmented Generation},
  booktitle    = {Findings of the Association for Computational Linguistics, {ACL} 2025,
                  Vienna, Austria, July 27 - August 1, 2025},
  series       = {Findings of {ACL}},
  pages        = {15506--15521},
  publisher    = {Association for Computational Linguistics},
  year         = {2025},
  url          = {https://aclanthology.org/2025.findings-acl.800/},
  bibsource    = {dblp computer science bibliography, https://dblp.org}
}

@inproceedings{cramming,
  author       = {Yuri Kuratov and
                  Mikhail Arkhipov and
                  Aydar Bulatov and
                  Mikhail Burtsev},
  editor       = {Wanxiang Che and
                  Joyce Nabende and
                  Ekaterina Shutova and
                  Mohammad Taher Pilehvar},
  title        = {Cramming 1568 Tokens into a Single Vector and Back Again: Exploring
                  the Limits of Embedding Space Capacity},
  booktitle    = {Proceedings of the 63rd Annual Meeting of the Association for Computational
                  Linguistics (Volume 1: Long Papers), {ACL} 2025, Vienna, Austria,
                  July 27 - August 1, 2025},
  pages        = {19323--19339},
  publisher    = {Association for Computational Linguistics},
  year         = {2025},
  url          = {https://aclanthology.org/2025.acl-long.948/},
  bibsource    = {dblp computer science bibliography, https://dblp.org}
}

@article{DBLP:journals/corr/abs-2312-10997,
  author       = {Yunfan Gao and
                  Yun Xiong and
                  Xinyu Gao and
                  Kangxiang Jia and
                  Jinliu Pan and
                  Yuxi Bi and
                  Yi Dai and
                  Jiawei Sun and
                  Qianyu Guo and
                  Meng Wang and
                  Haofen Wang},
  title        = {Retrieval-Augmented Generation for Large Language Models: {A} Survey},
  journal      = {CoRR},
  volume       = {abs/2312.10997},
  year         = {2023},
  url          = {https://doi.org/10.48550/arXiv.2312.10997},
  doi          = {10.48550/ARXIV.2312.10997},
  eprinttype    = {arXiv},
  eprint       = {2312.10997},
  bibsource    = {dblp computer science bibliography, https://dblp.org}
}

@inproceedings{DBLP:conf/acl/MinSL0YHZ23,
  author       = {Sewon Min and
                  Weijia Shi and
                  Mike Lewis and
                  Xilun Chen and
                  Wen{-}tau Yih and
                  Hannaneh Hajishirzi and
                  Luke Zettlemoyer},
  editor       = {Anna Rogers and
                  Jordan L. Boyd{-}Graber and
                  Naoaki Okazaki},
  title        = {Nonparametric Masked Language Modeling},
  booktitle    = {Findings of the Association for Computational Linguistics: {ACL} 2023,
                  Toronto, Canada, July 9-14, 2023},
  pages        = {2097--2118},
  publisher    = {Association for Computational Linguistics},
  year         = {2023},
  url          = {https://doi.org/10.18653/v1/2023.findings-acl.132},
  doi          = {10.18653/V1/2023.FINDINGS-ACL.132},
  bibsource    = {dblp computer science bibliography, https://dblp.org}
}

@inproceedings{DBLP:conf/icml/GuuLTPC20,
  author       = {Kelvin Guu and
                  Kenton Lee and
                  Zora Tung and
                  Panupong Pasupat and
                  Ming{-}Wei Chang},
  title        = {Retrieval Augmented Language Model Pre-Training},
  booktitle    = {Proceedings of the 37th International Conference on Machine Learning,
                  {ICML} 2020, 13-18 July 2020, Virtual Event},
  series       = {Proceedings of Machine Learning Research},
  volume       = {119},
  pages        = {3929--3938},
  publisher    = {{PMLR}},
  year         = {2020},
  url          = {http://proceedings.mlr.press/v119/guu20a.html},
  bibsource    = {dblp computer science bibliography, https://dblp.org}
}

@inproceedings{DBLP:conf/naacl/ShiMYS0LZY24,
  author       = {Weijia Shi and
                  Sewon Min and
                  Michihiro Yasunaga and
                  Minjoon Seo and
                  Richard James and
                  Mike Lewis and
                  Luke Zettlemoyer and
                  Wen{-}tau Yih},
  editor       = {Kevin Duh and
                  Helena G{\'{o}}mez{-}Adorno and
                  Steven Bethard},
  title        = {{REPLUG:} Retrieval-Augmented Black-Box Language Models},
  booktitle    = {Proceedings of the 2024 Conference of the North American Chapter of
                  the Association for Computational Linguistics: Human Language Technologies
                  (Volume 1: Long Papers), {NAACL} 2024, Mexico City, Mexico, June 16-21,
                  2024},
  pages        = {8371--8384},
  publisher    = {Association for Computational Linguistics},
  year         = {2024},
  url          = {https://doi.org/10.18653/v1/2024.naacl-long.463},
  doi          = {10.18653/V1/2024.NAACL-LONG.463},
  bibsource    = {dblp computer science bibliography, https://dblp.org}
}

@inproceedings{DBLP:conf/iclr/Lin0CSL00KSLZY24,
  author       = {Xi Victoria Lin and
                  Xilun Chen and
                  Mingda Chen and
                  Weijia Shi and
                  Maria Lomeli and
                  Richard James and
                  Pedro Rodriguez and
                  Jacob Kahn and
                  Gergely Szilvasy and
                  Mike Lewis and
                  Luke Zettlemoyer and
                  Wen{-}tau Yih},
  title        = {{RA-DIT:} Retrieval-Augmented Dual Instruction Tuning},
  booktitle    = {The Twelfth International Conference on Learning Representations,
                  {ICLR} 2024, Vienna, Austria, May 7-11, 2024},
  publisher    = {OpenReview.net},
  year         = {2024},
  url          = {https://openreview.net/forum?id=22OTbutug9},
  bibsource    = {dblp computer science bibliography, https://dblp.org}
}

@inproceedings{selfrag,
  author       = {Akari Asai and
                  Zeqiu Wu and
                  Yizhong Wang and
                  Avirup Sil and
                  Hannaneh Hajishirzi},
  title        = {{Self-RAG}: Learning to Retrieve, Generate, and Critique through Self-Reflection},
  booktitle    = {The Twelfth International Conference on Learning Representations,
                  {ICLR} 2024, Vienna, Austria, May 7-11, 2024},
  publisher    = {OpenReview.net},
  year         = {2024},
  url          = {https://openreview.net/forum?id=hSyW5go0v8},
  bibsource    = {dblp computer science bibliography, https://dblp.org}
}

@article{ccs,
  author       = {Zongqian Li and
                  Yinhong Liu and
                  Yixuan Su and
                  Nigel Collier},
  title        = {Prompt Compression for Large Language Models: {A} Survey},
  journal      = {CoRR},
  volume       = {abs/2410.12388},
  year         = {2024},
  url          = {https://doi.org/10.48550/arXiv.2410.12388},
  doi          = {10.48550/ARXIV.2410.12388},
  eprinttype    = {arXiv},
  eprint       = {2410.12388},
  bibsource    = {dblp computer science bibliography, https://dblp.org}
}

@inproceedings{dac,
  author       = {Yi Zhao and
                  Zuchao Li and
                  Hai Zhao and
                  Baoyuan Qi and
                  Guoming Liu},
  editor       = {Wanxiang Che and
                  Joyce Nabende and
                  Ekaterina Shutova and
                  Mohammad Taher Pilehvar},
  title        = {{DAC:} {A} Dynamic Attention-aware Approach for Task-Agnostic Prompt
                  Compression},
  booktitle    = {Proceedings of the 63rd Annual Meeting of the Association for Computational
                  Linguistics (Volume 1: Long Papers), {ACL} 2025, Vienna, Austria,
                  July 27 - August 1, 2025},
  pages        = {19395--19407},
  publisher    = {Association for Computational Linguistics},
  year         = {2025},
  url          = {https://aclanthology.org/2025.acl-long.952/},
  bibsource    = {dblp computer science bibliography, https://dblp.org}
}

@inproceedings{DBLP:conf/aaai/LiskavetsURKEL25,
  author       = {Barys Liskavets and
                  Maxim Ushakov and
                  Shuvendu Roy and
                  Mark Klibanov and
                  Ali Etemad and
                  Shane K. Luke},
  editor       = {Toby Walsh and
                  Julie Shah and
                  Zico Kolter},
  title        = {Prompt Compression with Context-Aware Sentence Encoding for Fast and
                  Improved {LLM} Inference},
  booktitle    = {Thirty-Ninth {AAAI} Conference on Artificial Intelligence, Thirty-Seventh
                  Conference on Innovative Applications of Artificial Intelligence,
                  Fifteenth Symposium on Educational Advances in Artificial Intelligence,
                  {AAAI} 2025, Philadelphia, PA, USA, February 25 - March 4, 2025},
  pages        = {24595--24604},
  publisher    = {{AAAI} Press},
  year         = {2025},
  url          = {https://doi.org/10.1609/aaai.v39i23.34639},
  doi          = {10.1609/AAAI.V39I23.34639},
  bibsource    = {dblp computer science bibliography, https://dblp.org}
}

@article{attentionrag,
  author       = {Yixiong Fang and
                  Tianran Sun and
                  Yuling Shi and
                  Xiaodong Gu},
  title        = {{AttentionRAG}: Attention-Guided Context Pruning in Retrieval-Augmented
                  Generation},
  journal      = {CoRR},
  volume       = {abs/2503.10720},
  year         = {2025},
  url          = {https://doi.org/10.48550/arXiv.2503.10720},
  doi          = {10.48550/ARXIV.2503.10720},
  eprinttype   = {arXiv},
  eprint       = {2503.10720},
  bibsource    = {dblp computer science bibliography, https://dblp.org}
}

@article{autoencoding,
  author       = {Xin Liu and
                  Runsong Zhao and
                  Pengcheng Huang and
                  Xinyu Liu and
                  Junyi Xiao and
                  Chunyang Xiao and
                  Tong Xiao and
                  Shengxiang Gao and
                  Zhengtao Yu and
                  Jingbo Zhu},
  title        = {Autoencoding-Free Context Compression for {LLMs} via Contextual Semantic
                  Anchors},
  journal      = {CoRR},
  volume       = {abs/2510.08907},
  year         = {2025},
  url          = {https://doi.org/10.48550/arXiv.2510.08907},
  doi          = {10.48550/ARXIV.2510.08907},
  eprinttype   = {arXiv},
  eprint       = {2510.08907},
  bibsource    = {dblp computer science bibliography, https://dblp.org}
}

@article{sara,
  author       = {Yiqiao Jin and
                  Kartik Sharma and
                  Vineeth Rakesh and
                  Yingtong Dou and
                  Menghai Pan and
                  Mahashweta Das and
                  Srijan Kumar},
  title        = {{SARA:} Selective and Adaptive Retrieval-augmented Generation with
                  Context Compression},
  journal      = {CoRR},
  volume       = {abs/2507.05633},
  year         = {2025},
  url          = {https://doi.org/10.48550/arXiv.2507.05633},
  doi          = {10.48550/ARXIV.2507.05633},
  eprinttype   = {arXiv},
  eprint       = {2507.05633},
  bibsource    = {dblp computer science bibliography, https://dblp.org}
}

@inproceedings{pcc,
  author       = {Yuhong Dai and
                  Jianxun Lian and
                  Yitian Huang and
                  Wei Zhang and
                  Mingyang Zhou and
                  Mingqi Wu and
                  Xing Xie and
                  Hao Liao},
  editor       = {Wanxiang Che and
                  Joyce Nabende and
                  Ekaterina Shutova and
                  Mohammad Taher Pilehvar},
  title        = {Pretraining Context Compressor for Large Language Models with Embedding-Based
                  Memory},
  booktitle    = {Proceedings of the 63rd Annual Meeting of the Association for Computational
                  Linguistics (Volume 1: Long Papers), {ACL} 2025, Vienna, Austria,
                  July 27 - August 1, 2025},
  pages        = {28715--28732},
  publisher    = {Association for Computational Linguistics},
  year         = {2025},
  url          = {https://aclanthology.org/2025.acl-long.1394/},
  bibsource    = {dblp computer science bibliography, https://dblp.org}
}

@inproceedings{discomp,
  author       = {Quancai Liu and
                  Haihui Fan and
                  Jinchao Zhang and
                  Xiangfang Li and
                  Chuanrong Li and
                  Bo Li},
  editor       = {Luis Chiruzzo and
                  Alan Ritter and
                  Lu Wang},
  title        = {DisComp: {A} Two-Stage Prompt Optimization Framework Combining Task-Agnostic
                  and Task-Aware Compression},
  booktitle    = {Findings of the Association for Computational Linguistics: {NAACL}
                  2025, Albuquerque, New Mexico, USA, April 29 - May 4, 2025},
  series       = {Findings of {ACL}},
  pages        = {1033--1044},
  publisher    = {Association for Computational Linguistics},
  year         = {2025},
  url          = {https://doi.org/10.18653/v1/2025.findings-naacl.58},
  doi          = {10.18653/V1/2025.FINDINGS-NAACL.58},
  bibsource    = {dblp computer science bibliography, https://dblp.org}
}

@article{OSCAR,
  author       = {Maxime Louis and
                  Thibault Formal and
                  Herv{\'{e}} D{\'{e}}jean and
                  St{\'{e}}phane Clinchant},
  title        = {{OSCAR:} Online Soft Compression And Reranking},
  journal      = {CoRR},
  volume       = {abs/2504.07109},
  year         = {2025},
  url          = {https://doi.org/10.48550/arXiv.2504.07109},
  doi          = {10.48550/ARXIV.2504.07109},
  eprinttype   = {arXiv},
  eprint       = {2504.07109},
  bibsource    = {dblp computer science bibliography, https://dblp.org}
}

@inproceedings{ACC-RAG,
  author       = {Shuyu Guo and
                  Shuo Zhang and
                  Zhaochun Ren},
  editor       = {Christos Christodoulopoulos and
                  Tanmoy Chakraborty and
                  Carolyn Rose and
                  Violet Peng},
  title        = {Enhancing {RAG} Efficiency with Adaptive Context Compression},
  booktitle    = {Findings of the Association for Computational Linguistics: {EMNLP}
                  2025, Suzhou, China, November 4-9, 2025},
  pages        = {24061--24076},
  publisher    = {Association for Computational Linguistics},
  year         = {2025},
  url          = {https://aclanthology.org/2025.findings-emnlp.1307/},
  bibsource    = {dblp computer science bibliography, https://dblp.org}
}

@article{DBLP:journals/corr/abs-2602-13980,
  author       = {Guojie Liu and
                  Yiqi Wang and
                  Yanfeng Yang and
                  Wenqi Fan and
                  Songlei Jian and
                  Jianfeng Zhang and
                  Jie Yu},
  title        = {Cognitive Chunking for Soft Prompts: Accelerating Compressor Learning
                  via Block-wise Causal Masking},
  journal      = {CoRR},
  volume       = {abs/2602.13980},
  year         = {2026},
  url          = {https://doi.org/10.48550/arXiv.2602.13980},
  doi          = {10.48550/ARXIV.2602.13980},
  eprinttype   = {arXiv},
  eprint       = {2602.13980},
  bibsource    = {dblp computer science bibliography, https://dblp.org}
}

@inproceedings{DBLP:conf/emnlp/ZhaoLLHXXZ25,
  author       = {Runsong Zhao and
                  Xin Liu and
                  Xinyu Liu and
                  Pengcheng Huang and
                  Chunyang Xiao and
                  Tong Xiao and
                  JingBo Zhu},
  editor       = {Christos Christodoulopoulos and
                  Tanmoy Chakraborty and
                  Carolyn Rose and
                  Violet Peng},
  title        = {Position IDs Matter: An Enhanced Position Layout for Efficient Context
                  Compression in Large Language Models},
  booktitle    = {Findings of the Association for Computational Linguistics: {EMNLP}
                  2025, Suzhou, China, November 4-9, 2025},
  pages        = {17715--17734},
  publisher    = {Association for Computational Linguistics},
  year         = {2025},
  url          = {https://aclanthology.org/2025.findings-emnlp.962/},
  bibsource    = {dblp computer science bibliography, https://dblp.org}
}

@inproceedings{LLMLingua,
  author       = {Huiqiang Jiang and
                  Qianhui Wu and
                  Chin{-}Yew Lin and
                  Yuqing Yang and
                  Lili Qiu},
  editor       = {Houda Bouamor and
                  Juan Pino and
                  Kalika Bali},
  title        = {{LLMLingua}: Compressing Prompts for Accelerated Inference of Large
                  Language Models},
  booktitle    = {Proceedings of the 2023 Conference on Empirical Methods in Natural
                  Language Processing, {EMNLP} 2023, Singapore, December 6-10, 2023},
  pages        = {13358--13376},
  publisher    = {Association for Computational Linguistics},
  year         = {2023},
  url          = {https://doi.org/10.18653/v1/2023.emnlp-main.825},
  doi          = {10.18653/V1/2023.EMNLP-MAIN.825},
  bibsource    = {dblp computer science bibliography, https://dblp.org}
}

@inproceedings{recomp,
  author       = {Fangyuan Xu and
                  Weijia Shi and
                  Eunsol Choi},
  title        = {{RECOMP:} Improving Retrieval-Augmented LMs with Context Compression
                  and Selective Augmentation},
  booktitle    = {The Twelfth International Conference on Learning Representations,
                  {ICLR} 2024, Vienna, Austria, May 7-11, 2024},
  publisher    = {OpenReview.net},
  year         = {2024},
  url          = {https://openreview.net/forum?id=mlJLVigNHp},
  bibsource    = {dblp computer science bibliography, https://dblp.org}
}

@inproceedings{Provence,
  author       = {Nadezhda Chirkova and
                  Thibault Formal and
                  Vassilina Nikoulina and
                  St{\'{e}}phane Clinchant},
  title        = {Provence: efficient and robust context pruning for retrieval-augmented
                  generation},
  booktitle    = {The Thirteenth International Conference on Learning Representations,
                  {ICLR} 2025, Singapore, April 24-28, 2025},
  publisher    = {OpenReview.net},
  year         = {2025},
  url          = {https://openreview.net/forum?id=TDy5Ih78b4},
  bibsource    = {dblp computer science bibliography, https://dblp.org}
}

@article{DBLP:journals/corr/abs-2304-12102,
  author       = {Yucheng Li},
  title        = {Unlocking Context Constraints of {LLMs}: Enhancing Context Efficiency
                  of {LLMs} with Self-Information-Based Content Filtering},
  journal      = {CoRR},
  volume       = {abs/2304.12102},
  year         = {2023},
  url          = {https://doi.org/10.48550/arXiv.2304.12102},
  doi          = {10.48550/ARXIV.2304.12102},
  eprinttype   = {arXiv},
  eprint       = {2304.12102},
  bibsource    = {dblp computer science bibliography, https://dblp.org}
}

@inproceedings{DBLP:conf/acl/JinLDZZWLQD25,
  author       = {Jiajie Jin and
                  Xiaoxi Li and
                  Guanting Dong and
                  Yuyao Zhang and
                  Yutao Zhu and
                  Yongkang Wu and
                  Zhonghua Li and
                  Ye Qi and
                  Zhicheng Dou},
  editor       = {Wanxiang Che and
                  Joyce Nabende and
                  Ekaterina Shutova and
                  Mohammad Taher Pilehvar},
  title        = {Hierarchical Document Refinement for Long-context Retrieval-augmented
                  Generation},
  booktitle    = {Proceedings of the 63rd Annual Meeting of the Association for Computational
                  Linguistics (Volume 1: Long Papers), {ACL} 2025, Vienna, Austria,
                  July 27 - August 1, 2025},
  pages        = {3502--3520},
  publisher    = {Association for Computational Linguistics},
  year         = {2025},
  url          = {https://aclanthology.org/2025.acl-long.176/},
  bibsource    = {dblp computer science bibliography, https://dblp.org}
}

@article{CORE,
  author       = {Ziqiang Cui and
                  Yunpeng Weng and
                  Xing Tang and
                  Peiyang Liu and
                  Shiwei Li and
                  Bowei He and
                  Jiamin Chen and
                  Xiuqiang He and
                  Chen Ma},
  title        = {{CORE:} Lossless Compression for Retrieval-Augmented {LLMs} via Reinforcement
                  Learning},
  journal      = {CoRR},
  volume       = {abs/2508.19282},
  year         = {2025},
  url          = {https://doi.org/10.48550/arXiv.2508.19282},
  doi          = {10.48550/ARXIV.2508.19282},
  eprinttype   = {arXiv},
  eprint       = {2508.19282},
  bibsource    = {dblp computer science bibliography, https://dblp.org}
}

@inproceedings{EXIT,
  author       = {Taeho Hwang and
                  Sukmin Cho and
                  Soyeong Jeong and
                  Hoyun Song and
                  SeungYoon Han and
                  Jong C. Park},
  editor       = {Wanxiang Che and
                  Joyce Nabende and
                  Ekaterina Shutova and
                  Mohammad Taher Pilehvar},
  title        = {{EXIT:} Context-Aware Extractive Compression for Enhancing Retrieval-Augmented
                  Generation},
  booktitle    = {Findings of the Association for Computational Linguistics, {ACL} 2025,
                  Vienna, Austria, July 27 - August 1, 2025},
  series       = {Findings of {ACL}},
  pages        = {4895--4924},
  publisher    = {Association for Computational Linguistics},
  year         = {2025},
  url          = {https://aclanthology.org/2025.findings-acl.253/},
  bibsource    = {dblp computer science bibliography, https://dblp.org}
}

@inproceedings{compact,
  author       = {Chanwoong Yoon and
                  Taewhoo Lee and
                  Hyeon Hwang and
                  Minbyul Jeong and
                  Jaewoo Kang},
  editor       = {Yaser Al{-}Onaizan and
                  Mohit Bansal and
                  Yun{-}Nung Chen},
  title        = {CompAct: Compressing Retrieved Documents Actively for Question Answering},
  booktitle    = {Proceedings of the 2024 Conference on Empirical Methods in Natural
                  Language Processing, {EMNLP} 2024, Miami, FL, USA, November 12-16,
                  2024},
  pages        = {21424--21439},
  publisher    = {Association for Computational Linguistics},
  year         = {2024},
  url          = {https://doi.org/10.18653/v1/2024.emnlp-main.1194},
  doi          = {10.18653/V1/2024.EMNLP-MAIN.1194},
  bibsource    = {dblp computer science bibliography, https://dblp.org}
}

@article{filc,
  author       = {Zhiruo Wang and
                  Jun Araki and
                  Zhengbao Jiang and
                  Md. Rizwan Parvez and
                  Graham Neubig},
  title        = {Learning to Filter Context for Retrieval-Augmented Generation},
  journal      = {CoRR},
  volume       = {abs/2311.08377},
  year         = {2023},
  url          = {https://doi.org/10.48550/arXiv.2311.08377},
  doi          = {10.48550/ARXIV.2311.08377},
  eprinttype    = {arXiv},
  eprint       = {2311.08377},
  bibsource    = {dblp computer science bibliography, https://dblp.org}
}

@inproceedings{gist,
  author       = {Jesse Mu and
                  Xiang Li and
                  Noah D. Goodman},
  editor       = {Alice Oh and
                  Tristan Naumann and
                  Amir Globerson and
                  Kate Saenko and
                  Moritz Hardt and
                  Sergey Levine},
  title        = {Learning to Compress Prompts with Gist Tokens},
  booktitle    = {Advances in Neural Information Processing Systems 36: Annual Conference
                  on Neural Information Processing Systems 2023, NeurIPS 2023, New Orleans,
                  LA, USA, December 10 - 16, 2023},
  year         = {2023},
  url          = {http://papers.nips.cc/paper\_files/paper/2023/hash/3d77c6dcc7f143aa2154e7f4d5e22d68-Abstract-Conference.html},
  bibsource    = {dblp computer science bibliography, https://dblp.org}
}

@inproceedings{rmt,
  author       = {Aydar Bulatov and
                  Yuri Kuratov and
                  Mikhail Burtsev},
  editor       = {Sanmi Koyejo and
                  S. Mohamed and
                  A. Agarwal and
                  Danielle Belgrave and
                  K. Cho and
                  A. Oh},
  title        = {Recurrent Memory Transformer},
  booktitle    = {Advances in Neural Information Processing Systems 35: Annual Conference
                  on Neural Information Processing Systems 2022, NeurIPS 2022, New Orleans,
                  LA, USA, November 28 - December 9, 2022},
  year         = {2022},
  url          = {http://papers.nips.cc/paper\_files/paper/2022/hash/47e288629a6996a17ce50b90a056a0e1-Abstract-Conference.html},
  bibsource    = {dblp computer science bibliography, https://dblp.org}
}

@inproceedings{DAST,
  author       = {Shaoshen Chen and
                  Yangning Li and
                  Zishan Xu and
                  Yongqin Zeng and
                  Shunlong Wu and
                  Xinshuo Hu and
                  Zifei Shan and
                  Xin Su and
                  Jiwei Tang and
                  Yinghui Li and
                  Hai{-}Tao Zheng},
  editor       = {Wanxiang Che and
                  Joyce Nabende and
                  Ekaterina Shutova and
                  Mohammad Taher Pilehvar},
  title        = {{DAST:} Context-Aware Compression in {LLMs} via Dynamic Allocation of
                  Soft Tokens},
  booktitle    = {Findings of the Association for Computational Linguistics, {ACL} 2025,
                  Vienna, Austria, July 27 - August 1, 2025},
  series       = {Findings of {ACL}},
  pages        = {20544--20552},
  publisher    = {Association for Computational Linguistics},
  year         = {2025},
  url          = {https://aclanthology.org/2025.findings-acl.1055/},
  bibsource    = {dblp computer science bibliography, https://dblp.org}
}

@inproceedings{LLoCO,
  author       = {Sijun Tan and
                  Xiuyu Li and
                  Shishir G. Patil and
                  Ziyang Wu and
                  Tianjun Zhang and
                  Kurt Keutzer and
                  Joseph Gonzalez and
                  Raluca A. Popa},
  editor       = {Yaser Al{-}Onaizan and
                  Mohit Bansal and
                  Yun{-}Nung Chen},
  title        = {{LLoCO}: Learning Long Contexts Offline},
  booktitle    = {Proceedings of the 2024 Conference on Empirical Methods in Natural
                  Language Processing, {EMNLP} 2024, Miami, FL, USA, November 12-16,
                  2024},
  pages        = {17605--17621},
  publisher    = {Association for Computational Linguistics},
  year         = {2024},
  url          = {https://doi.org/10.18653/v1/2024.emnlp-main.975},
  doi          = {10.18653/V1/2024.EMNLP-MAIN.975},
  bibsource    = {dblp computer science bibliography, https://dblp.org}
}

@inproceedings{Beacon,
  author       = {Peitian Zhang and
                  Zheng Liu and
                  Shitao Xiao and
                  Ninglu Shao and
                  Qiwei Ye and
                  Zhicheng Dou},
  title        = {Long Context Compression with Activation Beacon},
  booktitle    = {The Thirteenth International Conference on Learning Representations,
                  {ICLR} 2025, Singapore, April 24-28, 2025},
  publisher    = {OpenReview.net},
  year         = {2025},
  url          = {https://openreview.net/forum?id=1eQT9OzfNQ},
  bibsource    = {dblp computer science bibliography, https://dblp.org}
}

@inproceedings{dodo,
  author       = {Guanghui Qin and
                  Corby Rosset and
                  Ethan C. Chau and
                  Nikhil Rao and
                  Benjamin Van Durme},
  editor       = {Lun{-}Wei Ku and
                  Andre Martins and
                  Vivek Srikumar},
  title        = {Dodo: Dynamic Contextual Compression for Decoder-only {LMs}},
  booktitle    = {Proceedings of the 62nd Annual Meeting of the Association for Computational
                  Linguistics (Volume 1: Long Papers), {ACL} 2024, Bangkok, Thailand,
                  August 11-16, 2024},
  pages        = {9961--9975},
  publisher    = {Association for Computational Linguistics},
  year         = {2024},
  url          = {https://doi.org/10.18653/v1/2024.acl-long.536},
  doi          = {10.18653/V1/2024.ACL-LONG.536},
  bibsource    = {dblp computer science bibliography, https://dblp.org}
}

@inproceedings{ftr,
  author       = {David Wingate and
                  Mohammad Shoeybi and
                  Taylor Sorensen},
  editor       = {Yoav Goldberg and
                  Zornitsa Kozareva and
                  Yue Zhang},
  title        = {Prompt Compression and Contrastive Conditioning for Controllability
                  and Toxicity Reduction in Language Models},
  booktitle    = {Findings of the Association for Computational Linguistics: {EMNLP}
                  2022, Abu Dhabi, United Arab Emirates, December 7-11, 2022},
  pages        = {5621--5634},
  publisher    = {Association for Computational Linguistics},
  year         = {2022},
  url          = {https://doi.org/10.18653/v1/2022.findings-emnlp.412},
  doi          = {10.18653/V1/2022.FINDINGS-EMNLP.412},
  bibsource    = {dblp computer science bibliography, https://dblp.org}
}

@inproceedings{autocompr,
  author       = {Alexis Chevalier and
                  Alexander Wettig and
                  Anirudh Ajith and
                  Danqi Chen},
  editor       = {Houda Bouamor and
                  Juan Pino and
                  Kalika Bali},
  title        = {Adapting Language Models to Compress Contexts},
  booktitle    = {Proceedings of the 2023 Conference on Empirical Methods in Natural
                  Language Processing, {EMNLP} 2023, Singapore, December 6-10, 2023},
  pages        = {3829--3846},
  publisher    = {Association for Computational Linguistics},
  year         = {2023},
  url          = {https://doi.org/10.18653/v1/2023.emnlp-main.232},
  doi          = {10.18653/V1/2023.EMNLP-MAIN.232},
  bibsource    = {dblp computer science bibliography, https://dblp.org}
}

@inproceedings{icae,
  author       = {Tao Ge and
                  Jing Hu and
                  Lei Wang and
                  Xun Wang and
                  Si{-}Qing Chen and
                  Furu Wei},
  title        = {In-context Autoencoder for Context Compression in a Large Language
                  Model},
  booktitle    = {The Twelfth International Conference on Learning Representations,
                  {ICLR} 2024, Vienna, Austria, May 7-11, 2024},
  publisher    = {OpenReview.net},
  year         = {2024},
  url          = {https://openreview.net/forum?id=uREj4ZuGJE},
  bibsource    = {dblp computer science bibliography, https://dblp.org}
}

@article{cocom,
  author       = {David Rau and
                  Shuai Wang and
                  Herv{\'{e}} D{\'{e}}jean and
                  St{\'{e}}phane Clinchant},
  title        = {Context Embeddings for Efficient Answer Generation in {RAG}},
  journal      = {CoRR},
  volume       = {abs/2407.09252},
  year         = {2024},
  url          = {https://doi.org/10.48550/arXiv.2407.09252},
  doi          = {10.48550/ARXIV.2407.09252},
  eprinttype    = {arXiv},
  eprint       = {2407.09252},
  bibsource    = {dblp computer science bibliography, https://dblp.org}
}

@inproceedings{bergen,
  author       = {David Rau and
                  Herv{\'{e}} D{\'{e}}jean and
                  Nadezhda Chirkova and
                  Thibault Formal and
                  Shuai Wang and
                  St{\'{e}}phane Clinchant and
                  Vassilina Nikoulina},
  editor       = {Yaser Al{-}Onaizan and
                  Mohit Bansal and
                  Yun{-}Nung Chen},
  title        = {{BERGEN:} {A} Benchmarking Library for Retrieval-Augmented Generation},
  booktitle    = {Findings of the Association for Computational Linguistics: {EMNLP}
                  2024, Miami, Florida, USA, November 12-16, 2024},
  series       = {Findings of {ACL}},
  pages        = {7640--7663},
  publisher    = {Association for Computational Linguistics},
  year         = {2024},
  url          = {https://doi.org/10.18653/v1/2024.findings-emnlp.449},
  doi          = {10.18653/V1/2024.FINDINGS-EMNLP.449},
  bibsource    = {dblp computer science bibliography, https://dblp.org}
}

@inproceedings{DeBERTa,
  author       = {Pengcheng He and
                  Jianfeng Gao and
                  Weizhu Chen},
  title        = {{DeBERTaV3}: Improving {DeBERTa} using {ELECTRA-Style} Pre-Training with
                  Gradient-Disentangled Embedding Sharing},
  booktitle    = {The Eleventh International Conference on Learning Representations,
                  {ICLR} 2023, Kigali, Rwanda, May 1-5, 2023},
  publisher    = {OpenReview.net},
  year         = {2023},
  url          = {https://openreview.net/forum?id=sE7-XhLxHA},
  bibsource    = {dblp computer science bibliography, https://dblp.org}
}

@article{splade,
  author       = {Carlos Lassance and
                  Herv{\'{e}} D{\'{e}}jean and
                  Thibault Formal and
                  St{\'{e}}phane Clinchant},
  title        = {SPLADE-v3: New baselines for {SPLADE}},
  journal      = {CoRR},
  volume       = {abs/2403.06789},
  year         = {2024},
  url          = {https://doi.org/10.48550/arXiv.2403.06789},
  doi          = {10.48550/ARXIV.2403.06789},
  eprinttype   = {arXiv},
  eprint       = {2403.06789},
  bibsource    = {dblp computer science bibliography, https://dblp.org}
}

@inproceedings{llmlingua2,
  author       = {Zhuoshi Pan and
                  Qianhui Wu and
                  Huiqiang Jiang and
                  Menglin Xia and
                  Xufang Luo and
                  Jue Zhang and
                  Qingwei Lin and
                  Victor R{\"{u}}hle and
                  Yuqing Yang and
                  Chin{-}Yew Lin and
                  H. Vicky Zhao and
                  Lili Qiu and
                  Dongmei Zhang},
  editor       = {Lun{-}Wei Ku and
                  Andre Martins and
                  Vivek Srikumar},
  title        = {{LLMLingua-2}: Data Distillation for Efficient and Faithful Task-Agnostic
                  Prompt Compression},
  booktitle    = {Findings of the Association for Computational Linguistics, {ACL} 2024,
                  Bangkok, Thailand and virtual meeting, August 11-16, 2024},
  series       = {Findings of {ACL}},
  pages        = {963--981},
  publisher    = {Association for Computational Linguistics},
  year         = {2024},
  url          = {https://doi.org/10.18653/v1/2024.findings-acl.57},
  doi          = {10.18653/V1/2024.FINDINGS-ACL.57},
  bibsource    = {dblp computer science bibliography, https://dblp.org}
}

@inproceedings{popqa,
  author       = {Alex Mallen and
                  Akari Asai and
                  Victor Zhong and
                  Rajarshi Das and
                  Daniel Khashabi and
                  Hannaneh Hajishirzi},
  editor       = {Anna Rogers and
                  Jordan L. Boyd{-}Graber and
                  Naoaki Okazaki},
  title        = {When Not to Trust Language Models: Investigating Effectiveness of
                  Parametric and Non-Parametric Memories},
  booktitle    = {Proceedings of the 61st Annual Meeting of the Association for Computational
                  Linguistics (Volume 1: Long Papers), {ACL} 2023, Toronto, Canada,
                  July 9-14, 2023},
  pages        = {9802--9822},
  publisher    = {Association for Computational Linguistics},
  year         = {2023},
  url          = {https://doi.org/10.18653/v1/2023.acl-long.546},
  doi          = {10.18653/V1/2023.ACL-LONG.546},
  bibsource    = {dblp computer science bibliography, https://dblp.org}
}

@inproceedings{asqa,
  author       = {Ivan Stelmakh and
                  Yi Luan and
                  Bhuwan Dhingra and
                  Ming{-}Wei Chang},
  editor       = {Yoav Goldberg and
                  Zornitsa Kozareva and
                  Yue Zhang},
  title        = {{ASQA:} Factoid Questions Meet Long-Form Answers},
  booktitle    = {Proceedings of the 2022 Conference on Empirical Methods in Natural
                  Language Processing, {EMNLP} 2022, Abu Dhabi, United Arab Emirates,
                  December 7-11, 2022},
  pages        = {8273--8288},
  publisher    = {Association for Computational Linguistics},
  year         = {2022},
  url          = {https://doi.org/10.18653/v1/2022.emnlp-main.566},
  doi          = {10.18653/V1/2022.EMNLP-MAIN.566},
  bibsource    = {dblp computer science bibliography, https://dblp.org}
}

@article{nq,
  author       = {Tom Kwiatkowski and
                  Jennimaria Palomaki and
                  Olivia Redfield and
                  Michael Collins and
                  Ankur P. Parikh and
                  Chris Alberti and
                  Danielle Epstein and
                  Illia Polosukhin and
                  Jacob Devlin and
                  Kenton Lee and
                  Kristina Toutanova and
                  Llion Jones and
                  Matthew Kelcey and
                  Ming{-}Wei Chang and
                  Andrew M. Dai and
                  Jakob Uszkoreit and
                  Quoc Le and
                  Slav Petrov},
  title        = {Natural Questions: a Benchmark for Question Answering Research},
  journal      = {Trans. Assoc. Comput. Linguistics},
  volume       = {7},
  pages        = {452--466},
  year         = {2019},
  url          = {https://doi.org/10.1162/tacl\_a\_00276},
  doi          = {10.1162/TACL\_A\_00276},
  bibsource    = {dblp computer science bibliography, https://dblp.org}
}

@inproceedings{trivalqa,
  author       = {Mandar Joshi and
                  Eunsol Choi and
                  Daniel S. Weld and
                  Luke Zettlemoyer},
  editor       = {Regina Barzilay and
                  Min{-}Yen Kan},
  title        = {{TriviaQA}: {A} Large Scale Distantly Supervised Challenge Dataset for
                  Reading Comprehension},
  booktitle    = {Proceedings of the 55th Annual Meeting of the Association for Computational
                  Linguistics, {ACL} 2017, Vancouver, Canada, July 30 - August 4, Volume
                  1: Long Papers},
  pages        = {1601--1611},
  publisher    = {Association for Computational Linguistics},
  year         = {2017},
  url          = {https://doi.org/10.18653/v1/P17-1147},
  doi          = {10.18653/V1/P17-1147},
  bibsource    = {dblp computer science bibliography, https://dblp.org}
}

@article{ref,
  author       = {Ronald J. Williams},
  title        = {Simple Statistical Gradient-Following Algorithms for Connectionist
                  Reinforcement Learning},
  journal      = {Mach. Learn.},
  volume       = {8},
  pages        = {229--256},
  year         = {1992},
  url          = {https://doi.org/10.1007/BF00992696},
  doi          = {10.1007/BF00992696},
  bibsource    = {dblp computer science bibliography, https://dblp.org}
}

@article{mistral,
  author       = {Albert Q. Jiang and
                  Alexandre Sablayrolles and
                  Arthur Mensch and
                  Chris Bamford and
                  Devendra Singh Chaplot and
                  Diego de Las Casas and
                  Florian Bressand and
                  Gianna Lengyel and
                  Guillaume Lample and
                  Lucile Saulnier and
                  L{\'{e}}lio Renard Lavaud and
                  Marie{-}Anne Lachaux and
                  Pierre Stock and
                  Teven Le Scao and
                  Thibaut Lavril and
                  Thomas Wang and
                  Timoth{\'{e}}e Lacroix and
                  William El Sayed},
  title        = {Mistral {7B}},
  journal      = {CoRR},
  volume       = {abs/2310.06825},
  year         = {2023},
  url          = {https://doi.org/10.48550/arXiv.2310.06825},
  doi          = {10.48550/ARXIV.2310.06825},
  eprinttype    = {arXiv},
  eprint       = {2310.06825},
  bibsource    = {dblp computer science bibliography, https://dblp.org}
}

@inproceedings{lora,
  author       = {Edward J. Hu and
                  Yelong Shen and
                  Phillip Wallis and
                  Zeyuan Allen{-}Zhu and
                  Yuanzhi Li and
                  Shean Wang and
                  Lu Wang and
                  Weizhu Chen},
  title        = {{LoRA}: Low-Rank Adaptation of Large Language Models},
  booktitle    = {The Tenth International Conference on Learning Representations, {ICLR}
                  2022, Virtual Event, April 25-29, 2022},
  publisher    = {OpenReview.net},
  year         = {2022},
  url          = {https://openreview.net/forum?id=nZeVKeeFYf9},
  bibsource    = {dblp computer science bibliography, https://dblp.org}
}

@inproceedings{HotpotQA,
  author       = {Zhilin Yang and
                  Peng Qi and
                  Saizheng Zhang and
                  Yoshua Bengio and
                  William W. Cohen and
                  Ruslan Salakhutdinov and
                  Christopher D. Manning},
  editor       = {Ellen Riloff and
                  David Chiang and
                  Julia Hockenmaier and
                  Jun'ichi Tsujii},
  title        = {{HotpotQA}: {A} Dataset for Diverse, Explainable Multi-hop Question
                  Answering},
  booktitle    = {Proceedings of the 2018 Conference on Empirical Methods in Natural
                  Language Processing, Brussels, Belgium, October 31 - November 4, 2018},
  pages        = {2369--2380},
  publisher    = {Association for Computational Linguistics},
  year         = {2018},
  url          = {https://doi.org/10.18653/v1/d18-1259},
  doi          = {10.18653/V1/D18-1259},
  bibsource    = {dblp computer science bibliography, https://dblp.org}
}

@misc{logitlens,
  author       = {nostalgebraist},
  title        = {interpreting {GPT}: the logit lens},
  year         = {2020},
  month        = {Aug},
  howpublished = {\url{https://www.lesswrong.com/posts/AcKRB8wDpdaN6v6ru/interpreting-gpt-the-logit-lens}},
  note         = {LessWrong blog post. Accessed: 2026-05-02}
}

@inproceedings{Fever,
    author = {Thorne, James and Vlachos, Andreas and Christodoulopoulos, Christos and Mittal, Arpit},
    title = {{FEVER}: a Large-scale Dataset for Fact Extraction and {VERification}},
    booktitle = {NAACL-HLT},
    year = {2018}
}

@article{BioASQ,
  title={{BioASQ-QA}: A manually curated corpus for Biomedical Question Answering},
  author={Krithara, Anastasia and Nentidis, Anastasios and Bougiatiotis, Konstantinos and Paliouras, Georgios},
  journal={Scientific data},
  volume={10},
  number={1},
  pages={170},
  year={2023},
  publisher={Nature Publishing Group UK London}
}

@inproceedings{COVID-QA,
  title={{COVID-QA}: A question answering dataset for {COVID-19}},
  author={M{\"o}ller, Timo and Reina, Anthony and Jayakumar, Raghavan and Pietsch, Malte},
  booktitle={Proceedings of the 1st Workshop on NLP for COVID-19 at ACL 2020},
  year={2020}
}

@article{grpo,
  author       = {Zhihong Shao and
                  Peiyi Wang and
                  Qihao Zhu and
                  Runxin Xu and
                  Junxiao Song and
                  Mingchuan Zhang and
                  Y. K. Li and
                  Y. Wu and
                  Daya Guo},
  title        = {{DeepSeekMath}: Pushing the Limits of Mathematical Reasoning in Open
                  Language Models},
  journal      = {CoRR},
  volume       = {abs/2402.03300},
  year         = {2024},
  url          = {https://doi.org/10.48550/arXiv.2402.03300},
  doi          = {10.48550/ARXIV.2402.03300},
  eprinttype   = {arXiv},
  eprint       = {2402.03300},
  bibsource    = {dblp computer science bibliography, https://dblp.org}
}

@inproceedings{tang2026comi,
  title={Comi: Coarse-to-fine context compression via marginal information gain},
  author={Tang, Jiwei and Liu, Shilei and Zhang, Zhicheng and Yuan, Yujin and Zheng, Libin and Zheng, Bo and others},
  booktitle={International Conference on Learning Representations},
  volume={2026},
  pages={46648--46666},
  year={2026}
}

@inproceedings{wei2025instructrag,
  title={InstructRAG: Instructing retrieval-augmented generation via self-synthesized rationales},
  author={Wei, Zhepei and Chen, Wei-Lin and Meng, Yu},
  booktitle={International Conference on Learning Representations},
  volume={2025},
  pages={82731--82754},
  year={2025}
}

@article{gutierrez2025HippoRAG,
  title={From rag to memory: Non-parametric continual learning for large language models},
  author={Guti{\'e}rrez, Bernal Jim{\'e}nez and Shu, Yiheng and Qi, Weijian and Zhou, Sizhe and Su, Yu},
  journal={arXiv preprint arXiv:2502.14802},
  year={2025}
}

@inproceedings{guan2026deeprag,
  title={Deeprag: Thinking to retrieve step by step for large language models},
  author={Guan, Xinyan and Zeng, Jiali and Meng, Fandong and Xin, Chunlei and Lu, Yaojie and Lin, Hongyu and Han, Xianpei and Sun, Le and Zhou, Jie},
  booktitle={International Conference on Learning Representations},
  volume={2026},
  pages={117056--117071},
  year={2026}
}

@inproceedings{ting2026bridging,
  author       = {Yujan Ting and
                  Xu Tang and
                  Terrence Chen and
                  Weijing Huang},
  editor       = {Maria Liakata and
                  Viviane P. Moreira and
                  Jiajun Zhang and
                  David Jurgens},
  title        = {Bridging the Memorization-Utilization Gap: Near-Lossless Context Compression
                  via Reinforcement Learning},
  booktitle    = {Proceedings of the 64th Annual Meeting of the Association for Computational
                  Linguistics (Volume 1: Long Papers), {ACL} 2026, San Diego, California,
                  United States, July 2-7, 2026},
  pages        = {14949--14972},
  publisher    = {Association for Computational Linguistics},
  year         = {2026},
  url          = {https://doi.org/10.18653/v1/2026.acl-long.682},
  doi          = {10.18653/V1/2026.ACL-LONG.682},
  bibsource    = {dblp computer science bibliography, https://dblp.org}
}

@inproceedings{IterCOMP,
  author       = {JungMin Yun and
                  YoungBin Kim},
  editor       = {Maria Liakata and
                  Viviane P. Moreira and
                  Jiajun Zhang and
                  David Jurgens},
  title        = {IterCOMP: Reasoning-aware Adaptive Prompt Compression for Multi-hop
                  Question Answering},
  booktitle    = {Proceedings of the 64th Annual Meeting of the Association for Computational
                  Linguistics (Volume 1: Long Papers), {ACL} 2026, San Diego, California,
                  United States, July 2-7, 2026},
  pages        = {33827--33840},
  publisher    = {Association for Computational Linguistics},
  year         = {2026},
  url          = {https://doi.org/10.18653/v1/2026.acl-long.1559},
  doi          = {10.18653/V1/2026.ACL-LONG.1559},
  bibsource    = {dblp computer science bibliography, https://dblp.org}
}

@article{BeyondPos,
  author       = {Jiwei Tang and
                  Zhijing Huang and
                  Xinyu Zhang and
                  Chen Jason Zhang and
                  Jianxing Yu and
                  Libin Zheng and
                  Rui Meng and
                  Jian Yin},
  title        = {Beyond Position Bias: Shifting Context Compression from Position-Driven
                  to Semantic-Driven},
  journal      = {CoRR},
  volume       = {abs/2605.09463},
  year         = {2026},
  url          = {https://doi.org/10.48550/arXiv.2605.09463},
  doi          = {10.48550/ARXIV.2605.09463},
  eprinttype   = {arXiv},
  eprint       = {2605.09463},
  bibsource    = {dblp computer science bibliography, https://dblp.org}
}

@article{poc,
  author       = {Runsong Zhao and
                  Shilei Liu and
                  Jiwei Tang and
                  Langming Liu and
                  Haibin Chen and
                  Weidong Zhang and
                  Yujin Yuan and
                  Tong Xiao and
                  Jingbo Zhu and
                  Wenbo Su and
                  Bo Zheng},
  title        = {PoC: Performance-oriented Context Compression for Large Language Models
                  via Performance Prediction},
  journal      = {CoRR},
  volume       = {abs/2603.19733},
  year         = {2026},
  url          = {https://doi.org/10.48550/arXiv.2603.19733},
  doi          = {10.48550/ARXIV.2603.19733},
  eprinttype   = {arXiv},
  eprint       = {2603.19733},
  bibsource    = {dblp computer science bibliography, https://dblp.org}
}

@inproceedings{ReadAsHuman,
  author       = {Jiwei Tang and
                  Shilei Liu and
                  Zhicheng Zhang and
                  Qingsong Lv and
                  Runsong Zhao and
                  Tingwei Lu and
                  Langming Liu and
                  Haibin Chen and
                  Yujin Yuan and
                  Hai{-}Tao Zheng and
                  Wenbo Su and
                  Bo Zheng},
  editor       = {Maria Liakata and
                  Viviane P. Moreira and
                  Jiajun Zhang and
                  David Jurgens},
  title        = {Read As Human: Compressing Context via Parallelizable Close Reading
                  and Skimming},
  booktitle    = {Proceedings of the 64th Annual Meeting of the Association for Computational
                  Linguistics (Volume 1: Long Papers), {ACL} 2026, San Diego, California,
                  United States, July 2-7, 2026},
  pages        = {28393--28406},
  publisher    = {Association for Computational Linguistics},
  year         = {2026},
  url          = {https://doi.org/10.18653/v1/2026.acl-long.1309},
  doi          = {10.18653/V1/2026.ACL-LONG.1309},
  bibsource    = {dblp computer science bibliography, https://dblp.org}
}

@inproceedings{XProvence,
  author       = {Youssef Mohamed and
                  Mohamed Elhoseiny and
                  Thibault Formal and
                  Nadezhda Chirkova},
  editor       = {Ricardo Campos and
                  Adam Jatowt and
                  Yanyan Lan and
                  Mohammad Aliannejadi and
                  Christine Bauer and
                  Sean MacAvaney and
                  Avishek Anand and
                  Zhaochun Ren and
                  Suzan Verberne and
                  Nan Bai and
                  Masoud Mansoury},
  title        = {XProvence: Zero-Cost Multilingual Context Pruning for Retrieval-Augmented
                  Generation},
  booktitle    = {Advances in Information Retrieval - 48th European Conference on Information
                  Retrieval, {ECIR} 2026, Delft, The Netherlands, March 29 - April 2,
                  2026, Proceedings, Part {II}},
  series       = {Lecture Notes in Computer Science},
  volume       = {16484},
  pages        = {470--480},
  publisher    = {Springer},
  year         = {2026},
  url          = {https://doi.org/10.1007/978-3-032-21300-6\_38},
  doi          = {10.1007/978-3-032-21300-6\_38},
  bibsource    = {dblp computer science bibliography, https://dblp.org}
}

@inproceedings{Rethinking,
  author       = {Yunhao Liu and
                  Zian Jia and
                  Xinyu Gao and
                  Kanjun Xu and
                  Yun Xiong},
  editor       = {Hakim Hacid and
                  Yoelle Maarek and
                  Francesco Bonchi and
                  Ido Guy and
                  Emine Yilmaz},
  title        = {Rethinking Soft Compression in Retrieval-Augmented Generation: {A}
                  Query-Conditioned Selector Perspective},
  booktitle    = {Proceedings of the {ACM} Web Conference 2026, {WWW} 2026, Dubai, United
                  Arab Emirates, originally scheduled for April 13-17, 2026, rescheduled
                  for June 29 - July 3, 2026},
  pages        = {2441--2452},
  publisher    = {{ACM}},
  year         = {2026},
  url          = {https://doi.org/10.1145/3774904.3792700},
  doi          = {10.1145/3774904.3792700},
  bibsource    = {dblp computer science bibliography, https://dblp.org}
}

@inproceedings{AgentOCR,
  author       = {Lang Feng and
                  Fuchao Yang and
                  Feng Chen and
                  Xin Cheng and
                  Haiyang Xu and
                  Zhenglin Wan and
                  Ming Yan and
                  Bo An},
  editor       = {Maria Liakata and
                  Viviane P. Moreira and
                  Jiajun Zhang and
                  David Jurgens},
  title        = {AgentOCR: Reimagining Agent History via Optical Self-Compression},
  booktitle    = {Proceedings of the 64th Annual Meeting of the Association for Computational
                  Linguistics (Volume 1: Long Papers), {ACL} 2026, San Diego, California,
                  United States, July 2-7, 2026},
  pages        = {5067--5086},
  publisher    = {Association for Computational Linguistics},
  year         = {2026},
  url          = {https://doi.org/10.18653/v1/2026.acl-long.230},
  doi          = {10.18653/V1/2026.ACL-LONG.230},
  bibsource    = {dblp computer science bibliography, https://dblp.org}
}

@inproceedings{Understanding,
  author       = {Weronika Lajewska and
                  Momchil Hardalov and
                  Laura Aina and
                  Neha Anna John and
                  Hang Su and
                  Llu{\'{\i}}s M{\`{a}}rquez},
  editor       = {Christos Christodoulopoulos and
                  Tanmoy Chakraborty and
                  Carolyn Rose and
                  Violet Peng},
  title        = {Understanding and Improving Information Preservation in Prompt Compression
                  for LLMs},
  booktitle    = {Findings of the Association for Computational Linguistics: {EMNLP}
                  2025, Suzhou, China, November 4-9, 2025},
  pages        = {17520--17541},
  publisher    = {Association for Computational Linguistics},
  year         = {2025},
  url          = {https://doi.org/10.18653/v1/2025.findings-emnlp.949},
  doi          = {10.18653/V1/2025.FINDINGS-EMNLP.949},
  bibsource    = {dblp computer science bibliography, https://dblp.org}
}

@article{Simple,
  author       = {Yair Feldman and
                  Yoav Artzi},
  title        = {Simple Context Compression: Mean-Pooling and Multi-Ratio Training},
  journal      = {CoRR},
  volume       = {abs/2510.20797},
  year         = {2025},
  url          = {https://doi.org/10.48550/arXiv.2510.20797},
  doi          = {10.48550/ARXIV.2510.20797},
  eprinttype   = {arXiv},
  eprint       = {2510.20797},
  bibsource    = {dblp computer science bibliography, https://dblp.org}
}

@article{DynamicCompressing,
  author       = {Jinwu Hu and
                  Wei Zhang and
                  Yufeng Wang and
                  Yu Hu and
                  Bin Xiao and
                  Mingkui Tan and
                  Qing Du},
  title        = {Dynamic Compressing Prompts for Efficient Inference of Large Language
                  Models},
  journal      = {CoRR},
  volume       = {abs/2504.11004},
  year         = {2025},
  url          = {https://doi.org/10.48550/arXiv.2504.11004},
  doi          = {10.48550/ARXIV.2504.11004},
  eprinttype   = {arXiv},
  eprint       = {2504.11004},
  bibsource    = {dblp computer science bibliography, https://dblp.org}
}

@inproceedings{Leveraging,
  author       = {Yunlong Zhao and
                  Haoran Wu and
                  Bo Xu},
  editor       = {Toby Walsh and
                  Julie Shah and
                  Zico Kolter},
  title        = {Leveraging Attention to Effectively Compress Prompts for Long-Context
                  LLMs},
  booktitle    = {Thirty-Ninth {AAAI} Conference on Artificial Intelligence, Thirty-Seventh
                  Conference on Innovative Applications of Artificial Intelligence,
                  Fifteenth Symposium on Educational Advances in Artificial Intelligence,
                  {AAAI} 2025, Philadelphia, PA, USA, February 25 - March 4, 2025},
  pages        = {26048--26056},
  publisher    = {{AAAI} Press},
  year         = {2025},
  url          = {https://doi.org/10.1609/aaai.v39i24.34800},
  doi          = {10.1609/AAAI.V39I24.34800},
  bibsource    = {dblp computer science bibliography, https://dblp.org}
}

@inproceedings{EfficientPrompt,
  author       = {Weizhi Fei and
                  Xueyan Niu and
                  Guoqing Xie and
                  Yingqing Liu and
                  Bo Bai and
                  Wei Han},
  editor       = {Danielle Belgrave and
                  Cheng Zhang and
                  Laura N. Montoya and
                  Hsuan{-}Tien Lin and
                  Razvan Pascanu and
                  Piotr Koniusz and
                  Marzyeh Ghassemi and
                  Nancy Chen and
                  Iv{\'{a}}n Vladimir Meza Ru{\'{\i}}z and
                  Arturo Loaiza{-}Bonilla},
  title        = {Efficient Prompt Compression with Evaluator Heads for Long-Context
                  Transformer Inference},
  booktitle    = {Advances in Neural Information Processing Systems 38: Annual Conference
                  on Neural Information Processing Systems 2025, NeurIPS 2025, San Diego,
                  CA, USA, December 2-7, 2025 / Mexico City, Mexico, November 30 - December
                  5, 2025},
  year         = {2025},
  url          = {http://papers.nips.cc/paper\_files/paper/2025/hash/e356ed5f27885c79c7cb597bb1107c94-Abstract-Conference.html},
  bibsource    = {dblp computer science bibliography, https://dblp.org}
}

@inproceedings{YenG024Long-ContextLanguage,
  author       = {Howard Yen and
                  Tianyu Gao and
                  Danqi Chen},
  editor       = {Lun{-}Wei Ku and
                  Andre Martins and
                  Vivek Srikumar},
  title        = {Long-Context Language Modeling with Parallel Context Encoding},
  booktitle    = {Proceedings of the 62nd Annual Meeting of the Association for Computational
                  Linguistics (Volume 1: Long Papers), {ACL} 2024, Bangkok, Thailand,
                  August 11-16, 2024},
  pages        = {2588--2610},
  publisher    = {Association for Computational Linguistics},
  year         = {2024},
  url          = {https://doi.org/10.18653/v1/2024.acl-long.142},
  doi          = {10.18653/V1/2024.ACL-LONG.142},
  bibsource    = {dblp computer science bibliography, https://dblp.org}
}
\bibliographystyle{iclr2027_conference}

\appendix
\section{LLM-as-Judge Prompt}
\label{sec:llm_judge_prompt}

The following prompt is used to query the judge model for the LLM-as-judge accuracy metric reported in our experiments.

\begin{tcblisting}{
    listing only,
    breakable,
    colback=gray!8,
    colframe=gray!40,
    boxrule=0.4pt,
    arc=2pt,
    left=6pt, right=6pt, top=6pt, bottom=6pt,
    listing options={
        basicstyle=\ttfamily\small,
        breaklines=true,
        breakindent=0pt,
        columns=fullflexible,
        keepspaces=true,
        showstringspaces=false,
    }
}
You are an evaluation tool. Just answer by {{ Yes }} or {{ No }}. 

Given a question, a list of golden answers (any one of them is considered correct), and an AI-generated answer, judge whether the AI-generated answer is correct - it is correct if it matches ANY ONE of the golden answers. 

Question: {question}
Golden answers: {ground_truth}

Generated answer: {model_answer}
Response: {{
\end{tcblisting}

%%%%%%%%%%%%%%%%%%%%%%%%%%%%%%%%%%%%%%%%%%%%%%%%%%%%%%%%%%%%

\section{Pure Distillation Hyperparameters}
\label{sec:appendix_pd_params}

Pure Distillation (\S\ref{sec:pd}) is performed on the teacher-correct subset $\mathcal{S}^{+}$, where correctness is determined by Gemini 3 Flash as described in Section~\ref{sec:impl}.
The core hyperparameters are listed in Table~\ref{tab:pd_hyperparams}.

\begin{table}[h]
    \centering
    \caption{Hyperparameters for Pure Distillation.}
    \label{tab:pd_hyperparams}
    \begin{tabular}{lc}
        \toprule
        \textbf{Hyperparameter} & \textbf{Assignment} \\
        \midrule
        optimizer & AdamW \\
        learning rate & $1\!\times\!10^{-4}$ \\
        warmup ratio & $0.05$ \\
        weight decay & $0.1$ \\
        gradient clipping & $1.0$ \\
        epochs & $1$ \\
        per-device batch size & $2$ \\
        gradient accumulation steps & $1$ \\
        effective batch size & $16$ \\
        max response length & $128$ \\
        distributed setup & DeepSpeed (8 GPUs) \\
        random seed & $42$ \\
        \bottomrule
    \end{tabular}
\end{table}

\section{Hard Exploration Hyperparameters}
\label{sec:appendix_he_params}

Hard Exploration (\S\ref{sec:he}) initializes from the PD checkpoint and runs GRPO on the teacher-failed subset $\mathcal{S}^{-}$. 
The core hyperparameters are listed in Table~\ref{tab:he_hyperparams}.

\begin{table}[h]
    \centering
    \caption{Hyperparameters for Hard Exploration.}
    \label{tab:he_hyperparams}
    \begin{tabular}{lc}
        \toprule
        \textbf{Hyperparameter} & \textbf{Assignment} \\
        \midrule
        optimizer & AdamW \\
        learning rate & $5\!\times\!10^{-6}$ \\
        warmup ratio & $0.05$ \\
        weight decay & $0.01$ \\
        gradient clipping & $1.0$ \\
        gradient checkpointing & True \\
        epochs & $2$ \\
        training micro batch size & $8$ \\
        per-GPU policy batch size & $64$ \\
        gradient accumulation steps & $1$ \\
        GRPO group size $G$ & $16$ \\
        clipping range $\varepsilon$ & $0.2$ \\
        KL coefficient $\beta$ & $0.02$ \\
        reward function & binary CEM \\
        rollout temperature & $1$ \\
        rollout top-$p$ & $0.95$ \\
        max generation length & $32$ \\
        random seed & $42$ \\
        \bottomrule
    \end{tabular}
\end{table}

\section{Comparison with Standard Distillation}
\label{sec:pisco_embedding_comparison}

\begin{figure}[!htbp]
    \centering
    \includegraphics[width=\textwidth]{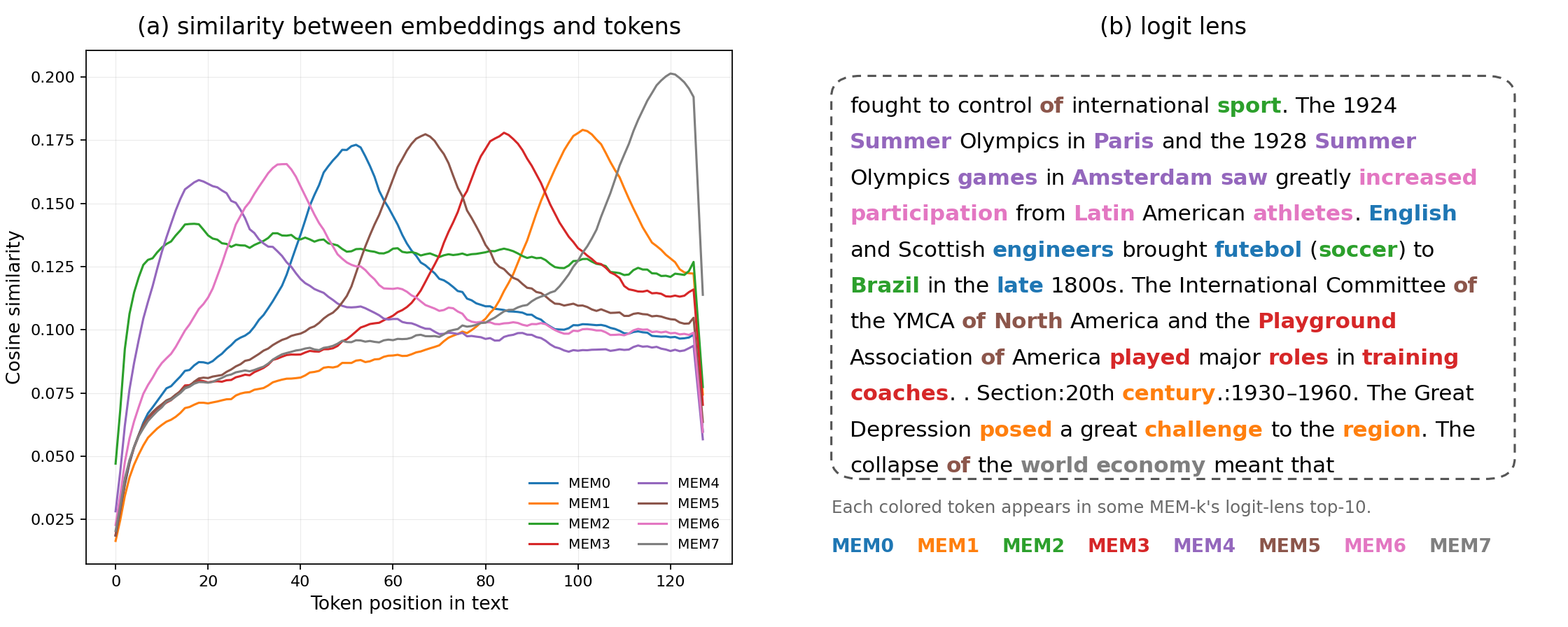}
    \caption{PISCO compression embeddings learned by standard distillation, for the same passage as Figure~\ref{fig:embeddings}. (a) Cosine similarity between embeddings and document tokens. (b) Document tokens appearing among the top-$10$ logit-lens vocabulary projections of the memory embeddings. Colors identify embeddings within each model.}
    \label{fig:embeddings_pisco}
\end{figure}

We compare the compression embeddings learned by PISCO through standard distillation (Figure~\ref{fig:embeddings_pisco}) with those of DEX-Comp (Figure~\ref{fig:embeddings}).
Both visualizations use the same passage and show eight memory embeddings.
We compare the distribution of information across each model's embeddings, whose indices are model-specific.

\paragraph{Global and local information.}
In this example, PISCO has one embedding with a broadly distributed similarity profile (MEM2), while the other seven exhibit localized peaks.
DEX-Comp has three embeddings (MEM2, MEM5, and MEM7) whose similarities remain relatively high across most of the document tokens, alongside five embeddings with localized peaks.
This pattern suggests that DEX-Comp allocates more embeddings to global information while retaining representations of specific text regions.

\paragraph{Logit-lens projections.}
The projections differ in both the highlighted vocabulary and its distribution across embeddings.
PISCO highlights the Olympic place names \emph{Paris} and \emph{Amsterdam}, which do not appear in DEX-Comp's displayed overlay.
DEX-Comp instead highlights \emph{major} in the coach-training passage.
Shared tokens are also distributed differently: \emph{increased participation} and \emph{Latin} appear under MEM6 in PISCO and MEM1 in DEX-Comp, while much of the coach-training passage appears under MEM3 in PISCO and MEM4 in DEX-Comp.
Together with the more broadly distributed similarity profiles, these differences reveal a distinct organization of compressed representations in this example.

\end{document}